# Zero and Few Shot Load Forecasting with Large Language Models

Wenlong Liao[a,b], Chengrui Zhang[c], Zhe Yang[d], Mengshuo Jia[e,f,g], Christian Rehtanz[h], Jiannong Fang[b], and Fernando Porté-Agel[b]

[a]School of Electrical Engineering, Southeast University, Nanjing 210096, China.
[b]Wind Engineering and Renewable Energy Laboratory, Ecole Polytechnique Federale de Lausanne (EPFL), Lausanne 1015, Switzerland.
[c]College of Electrical Engineering and New Energy, China Three Gorges University, Yichang 443002, China.
[d]Department of Electrical and Electronic Engineering, Imperial College London, London SW7 2AZ, United Kingdom.
[e]The Department of Automation, School of Automation and Intelligent Sensing, Shanghai Jiao Tong University, Shanghai 200240, China.
[f]The Key Laboratory of System Control and Information Processing, Ministry of Education of China, Shanghai 200240, China.
[g]State Key Laboratory of Submarine Geoscience, Shanghai 201913, China.
[h]Institute of Energy Systems, Energy Efficiency and Energy Economic, TU Dortmund University, 44227 Dortmund, Germany.

***Abstract***—Deep learning models have shown strong performance in load forecasting, but they generally require large amounts of data for model training before being applied to new scenarios, which limits their effectiveness in data-scarce scenarios. Inspired by the great success of pre-trained language models (LLMs) in natural language processing, this paper proposes a zero and few shot load forecasting approach using an advanced LLM framework denoted as the Chronos model. By utilizing its extensive pre-trained knowledge, the Chronos model enables accurate load forecasting in data-scarce scenarios. Simulation results across five real-world datasets demonstrate that the Chronos model significantly outperforms nine popular baseline models for both deterministic and probabilistic load forecasting with various forecast horizons (e.g., 1 to 48 hours), even though the Chronos model is neither tailored nor fine-tuned to these specific load datasets. Notably, Chronos reduces root mean squared error (RMSE), continuous ranked probability score (CRPS), and quantile score (QS) by approximately 7.34%-84.30%, 19.63%-60.06%, and 22.83%-54.49%, respectively, compared to baseline models. These results highlight the superiority and flexibility of the Chronos model, positioning it as an effective solution in data-scarce scenarios.

**Index Terms**—Load Forecasting, Large Language Model, Time series, Smart grid, Artificial intelligence

## 1 INTRODUCTION

Electric load forecasting plays a critical role in the efficient planning and management of power systems [1]. For example, short-term load forecasting significantly contributes to the optimal scheduling of generation units, reducing the reliance on excess reserve power. In contrast, medium-term and long-term load forecasting enable utilities to plan maintenance activities more effectively, ensuring the reliability and safety of grid operations [2]. To avoid ambiguity, we clarify that this paper focuses on electric load forecasting, with forecast horizons ranging from hours to days.

Traditionally, load forecasting with short forecast horizons (e.g., 30 minutes and 1 hour) has been dominated by various statistical models, such as the automatic error, trend, and seasonality model selection (AutoETS) [3], seasonal naive model (SNM) [4], croston syntetos and boylan approximate (CSBA) [5], and non-parametric time series (NPTS) [6]. For instance, the study in [7] utilizes the automatic autoregressive integrated moving average (AutoARIMA) for load forecasting across 709 households in Ireland. Similarly, the work in [8] applies the error, trend, and seasonality model to dynamically capture the key components of complex time series with seasonal patterns for short-term load forecasting in 35 European countries. Generally, statistical models provide high interpretability and low computational complexity while requiring a few training data. However, their accuracy significantly diminishes for longer forecast horizons (e.g., 24 hours and 48 hours) due to their inability to effectively capture complex nonlinear relationships [9].

The focus of load forecasting has shifted from traditional statistical models to machine learning models (particularly deep neural networks) recently. Prominent machine learning models used in load forecasting include graph neural networks, light gradient boosting machines, regression trees, extreme gradient boosting, recurrent neural networks, transformer architectures, and their hybrids or variants [10]. For example, the study in [11] employs the deep autoregressive (DeepAR) model to capture seasonal patterns based on given covariates across time series for building load forecasting. Similarly, the work in [12] utilizes the temporal fusion transformer (TFT) to model long-term temporal dependencies for probabilistic forecasting of medium-term loads. The patch time series transformer (PatchTST), as proposed in [13], addresses inter-feature dependencies and trend variations for net load forecasting. Despite their strong performance, these deep learning models are generally limited to training and forecasts within the same dataset, hindering their generalization to unseen data. In other words, they typically require model retraining before being applied to new scenarios or datasets, which can be both data-intensive and time-consuming. This limitation highlights the need for more flexible methods, such as zero-shot learning models, which can enable accurate load forecasting in data-scarce scenarios. By definition, zero-shot learning does not require extensive re-training on specific datasets, making it effective when no task-specific data is available.

In recent years, pre-trained language models (LLMs) have garnered increasing attention due to the advent of the chat generative pre-trained transformer (ChatGPT) [14]. By definition, these models are large deep neural networks trained on extensive and diverse datasets to develop a versatile framework that can be fine-tuned or directly applied to a wide range of downstream tasks. By now, LLMs have significantly advanced the development of zero-shot learning in different fields [15], such as natural language processing and computer vision. For example, the work in [16] introduces a visual LLM for zero-shot image classification, while [17] explores the use of popular LLMs for conversational recommendation tasks in zero-shot settings.

Inspired by the great success of pre-trained LLMs, a few pioneering works have attempted to extend their applications into general time series forecasting [18], but load forecasting remains largely unexplored. Specifically, a straightforward research line involves mapping time series data to text prompts, thereby transforming the forecasting task into a question-answering format for LLMs. For example, [19] directly applies LLMs to weather forecasting and human mobility forecasting by converting numerical inputs and outputs into prompts, while [20] introduces a reprogramming framework to preprocess the input time series before feeding them into the frozen LLM. Similarly, [21] constructs the zero-shot forecasting problem as a next-token prediction in LLMs (e.g., GPT-3 and LLaMA-2) by encoding real-valued data into strings. Another research line is to fine-tune pre-trained LLMs for time series forecasting. For example, [22] freezes most the pre-trained layers in LLMs and fine-tunes only the positional embeddings for different tasks. In [23], an autoregressive fine-tuning strategy combined with spatio-temporal prompts is proposed to adapt LLMs to wind power data.

Despite the promising potential of LLMs in time series forecasting, these methods face notable limitations, such as the need for prompt design or model fine-tuning for each specific forecasting task, which demands extensive computational resources and increases inference time. Moreover, the application of LLMs to load forecasting is rarely discussed. In other words, the potential of LLMs in load forecasting has not received enough attention.

To address these limitations, this paper proposes a zero and few shot load forecasting approach by adopting LLMs. In particular, this paper proposes to use the recently developed LLM [24], an advanced artificial intelligence framework denoted as the Chronos model, for load forecasting. The Chronos model, a family of pre-trained models based on LLM architectures, has emerged as a cutting-edge framework in general time series analysis, with the promise of utilizing diverse training data. Its ability to handle varied time series datasets makes it an ideal candidate for load forecasting, which is inherently a time series forecasting task. To this end, this paper explores how the Chronos model can be effectively applied to load forecasting, particularly in data-scarce scenarios.

The main contributions and innovations of this paper are as follows:

- **New perspective:** Most existing works rely on traditional machine learning and statistical models with simple structures and few parameters for load forecasting. In contrast, this paper extends LLMs with complex architectures and extensive parameters to load forecasting, which remains largely unexplored.

- **Novel Application:** This paper proposes a zero and few shot load forecasting approach by adopting advanced LLMs, specifically the Chronos model. To the best of our knowledge, this is the first work applying the Chronos model to load forecasting.

- **Efficient Algorithms:** We demonstrate that the Chronos model can directly forecast loads for new or unseen scenarios even without requiring prompt engineering for a specific load dataset. This capability is particularly beneficial for scenarios lacking historical load data, such as emerging regions or situations constrained by privacy concerns. Simulation results across five real-world datasets demonstrate that the Chronos model significantly outperforms nine popular baseline models for both deterministic and probabilistic load forecasting with various forecast horizons.

The rest is organized as follows: Section 2 rigorously formulates the load forecasting. Section 3 proposes and describes the Chronos model in load forecasting. Simulations and results are discussed in Section 4 and Section 5. Finally, Section 6 presents the conclusion.

## 2 Problem Formulation

This section presents the mathematical descriptions for two types of load forecasting tasks that we are interested in, including deterministic load forecasting and probabilistic load forecasting.

### 2.1 Deterministic Load Forecasting

The objective of deterministic load forecasting is to estimate future load values (i.e., deterministic data points) at multiple time points given a set of historical load data and numerical weather prediction (NWP) data. The mathematical formulation of deterministic load forecasting is as follows [1]:

$$\hat{x}_{t+1},\hat{x}_{t+2},\ldots,\hat{x}_{t+h} = f\left(x_t, x_{t-1},\ldots,x_{t-n};\mathrm{NWP}_{t+1},\mathrm{NWP}_{t+2},\ldots,\mathrm{NWP}_{t+h}\right) \tag{1}$$

where $\hat{x}_{t+1},\hat{x}_{t+2},\ldots,\hat{x}_{t+h}$ are the forecast values of loads from time $t$+1 to time $t$+$h$; $\mathrm{NWP}_{t+1}$,$\mathrm{NWP}_{t+2}$,…,$\mathrm{NWP}_{t+h}$ are the NWP values from time $t$+1 to time $t$+$h$; $x_t$,$x_{t-1}$,…,$x_{t-n}$ are the historical values of loads from time $t$-$n$ to time $t$; $h$ is the forecast horizon; $n$ is the length of the historical window used for load forecasting; and $f$ is a machine learning model.

### 2.2 Probabilistic Load Forecasting

Compared to deterministic load forecasting, probabilistic load forecasting not only provides point forecast values, but also represents load uncertainty. Here, this paper uses the quantile form for output, generating forecast values at different confidence levels for multiple future time points, given the historical load data and NWP data as input.

The key principle of quantile regression is to provide forecast values at different quantiles for each future time point, generating the load values at various confidence levels. For instance, the 10th percentile $\alpha$=0.1 means that there is a 10% probability that the load will be below the forecast value. The mathematical formulation of probabilistic load forecasting is as follows [2]:

$$\hat{x}^{\alpha}_{t+1},\hat{x}^{\alpha}_{t+2},\ldots,\hat{x}^{\alpha}_{t+h} = f\left(x_t, x_{t-1},\ldots,x_{t-n},\alpha;\mathrm{NWP}_{t+1},\mathrm{NWP}_{t+2},\ldots,\mathrm{NWP}_{t+h}\right) \tag{2}$$

where $\hat{x}^{\alpha}_{t+1},\hat{x}^{\alpha}_{t+2},\ldots,\hat{x}^{\alpha}_{t+h}$ are the forecast values of quantile regression from time $t$+1 to time $t$+$h$; and $\alpha$ is the specified quantile (e.g., $\alpha$=0.1,0.2,...,0.9).

## 3 Chronos Model in Load Forecasting

This section introduces the Chronos model and how it can be adapted to the load forecasting tasks that we are interested in. The following subsections detail specific aspects of the Chronos model: scaling and quantization, data augmentation, model structure, model training, model fine-tuning, forecasting with covariates and implementation details.

### 3.1 Scaling and Quantization

As we all know, LLMs are originally designed for natural language processing. Although both language and time series share a similar sequential structure, they differ in how they are represented. In particular, natural language is composed of text drawn from the vocabulary, whereas time series consist of continuous real values. This difference requires particular adjustments to the time series data, particularly with regard to tokenization, in order to effectively input them to LLMs. Specifically, time series data needs to be transformed into discrete tokens through scaling and quantization, which are then fed into LLMs for model training.

***1) Scaling***

Time series data may vary widely in scale, even within a single dataset, which can result in optimization difficulties for deep learning models [9]. To improve optimization, each individual time series is normalized to an appropriate range for quantization. There are many techniques for data scaling, such as standard scaling, min-max scaling, and mean scaling. Here, the mean scaling technique is chosen, as it has proven effective in deep learning models [10]. The mathematical formulation of mean scaling is as follows:

$$\tilde{x}_i = \left(x_t - m\right) / s \tag{3}$$

where $s$ is the average value of the absolute values of the time series; and $m$ is set to 0.

Fig. 1(a) and Fig.1(b) present an illustration of mean scaling applied to the Midea dataset in [25]. It suggests that scaling not only allows the data to retain its temporal shape, but also constrains the values to a more limited range. This approach simplifies the representation of these values as a fixed set of tokens for training the LLM.

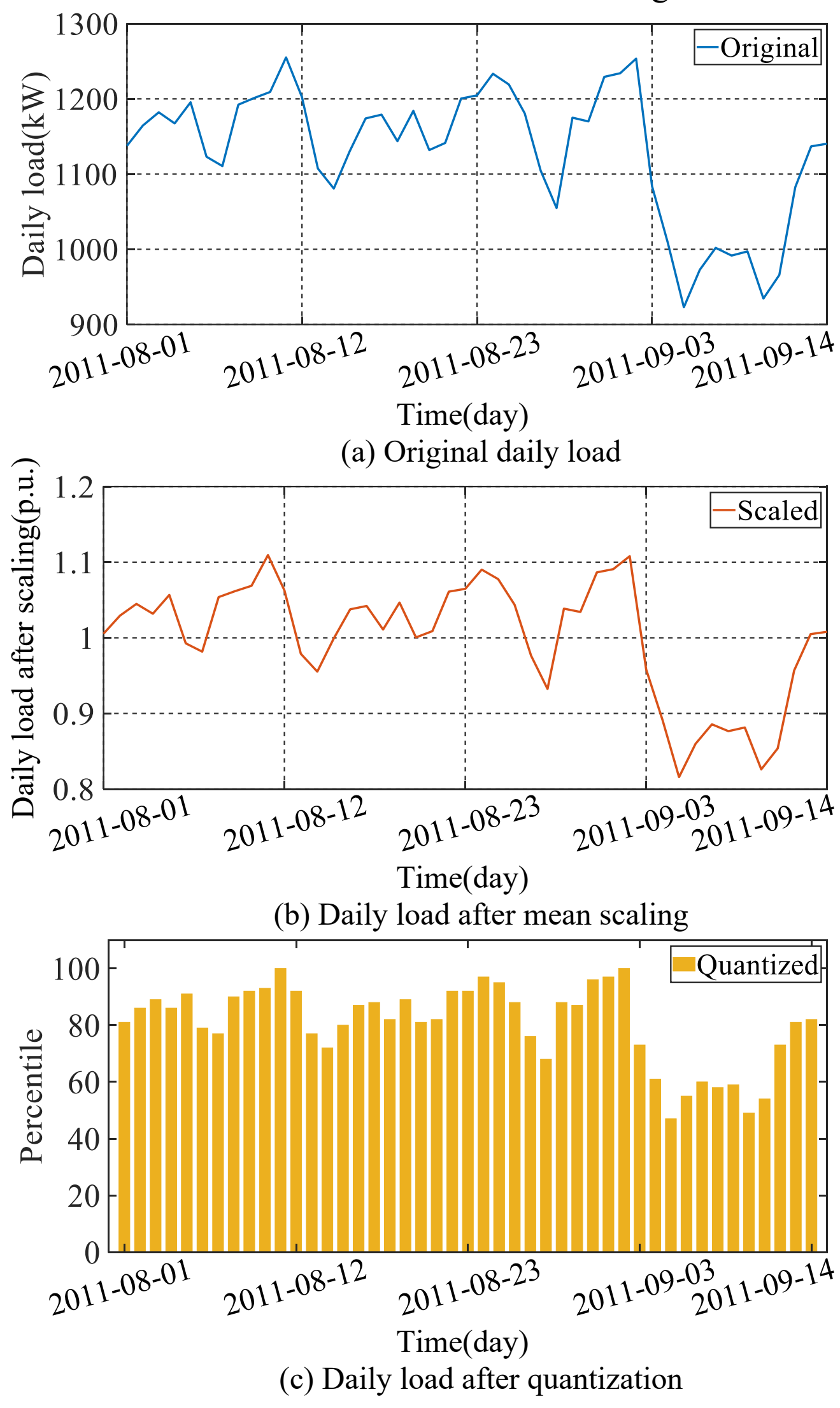

Fig. 1. A simple example of mean scaling and quantization on load data.

### 2) Quantization

After scaling the data, the next step is quantization. This is an essential process, because the scaled data remains continuous, which LLMs cannot process. Quantization transforms continuous real values into discrete values. This transformation allows us to generate context tokens from time series data for LLM input.

In particular, the widely used uniform binning technique is used to achieve data discretization. In this technique, bins are created such that the smallest value corresponds to the first bin and the largest value in the series is assigned to the last bin. Each bin edge is then evenly distributed across the range of values in the series, with each individual value assigned to a specific bin. The mathematical formulation of quantization is as follows:

$$g(x)=\begin{cases}1, & \text{if } x_{\min}\le x<x_{\min}+\Delta d\\ 2, & \text{if } x_{\min}+\Delta d\le x<x_{\min}+2\Delta d\\ \vdots & \\ N, & \text{if } x_{\min}+(N-1)\Delta d\le x<x_{\max}\end{cases} \tag{4}$$

where $\Delta d=(x_{\max}-x_{\min})/N$; $N$ is the number of the bin; $X_{\min}$ is the minimum continuous real value; and $X_{\min}$ is the maximum continuous real value.

Fig. 1(b) and Fig.1(c) present an illustration of quantization by percentile binning. Specifically, 100 bins are constructed, each corresponding to a percentile, and the scaled values are then assigned to their respective bins. In this case, the time series is transformed into a sequence of fixed and discrete tokens, as percentile binning restricts the token values to a range between 1 and 100. This process effectively tokenizes the time series, transforming it from continuous real values into a predefined set of tokens, which is the format required for LLMs.

Note that quantization inevitably introduces a certain level of information compression by mapping continuous values to discrete tokens. However, this design choice does not fundamentally limit forecast accuracy in the Chronos framework. First, the quantization resolution can be flexibly controlled by the number of bins, allowing a fine-grained approximation of the original continuous values. Second, Chronos formulates forecasting as a probabilistic token prediction task rather than direct point-wise regression, enabling the model to learn a distribution over quantized values, which effectively mitigates the impact of discretization noise. Moreover, the final forecasts are obtained by mapping predicted tokens back to continuous values, resulting in smooth and accurate predictions in practice. Empirical results in Sections 4 and 5 further demonstrate that, despite using quantization, Chronos consistently outperforms continuous regression-based baselines, indicating that the potential information loss introduced by uniform binning does not hinder its forecasting accuracy.

## 3.2 Data Augmentation

The training of LLMs relies on large and diverse datasets, which is relatively easy to collect in natural language processing due to the availability of many high-quality public text datasets. However, publicly available time series datasets are relatively scarce, which poses a challenge to model training of LLMs for time series forecasting.

To address this issue, data augmentation techniques are employed to enhance the diversity of the training data. Specifically, the widely used time series mixup technique is employed to randomly select sub-sequences from existing time series data, and then generate new training data by linear combinations of these sub-sequences. The mathematical formulation of the time series mixup technique is as follows:

$$x_{1:l}^{\text{new}}=\sum_{i=1}^{k}\lambda_i x_{1:l}^{(i)} \tag{5}$$

where $x_{1:l}^{\text{new}}$ is the generated time series data; $x_{1:l}^{(i)}$ the i-th selected time series; and $l$ is the number of data points in the time series data. Note that 2 and 3 sub-sequences are selected for linear combination here, as a smaller number of samples can more effectively preserve the unique patterns of each sample, thus avoiding excessive mixing of information that may lead to the loss of important time series features.

Fig. 2 presents a simple example of the time series mixup technique, which combines the selected two time series using different weights to create a new time series for model training.

Fig. 2. A simple example of the time series mixup technique.

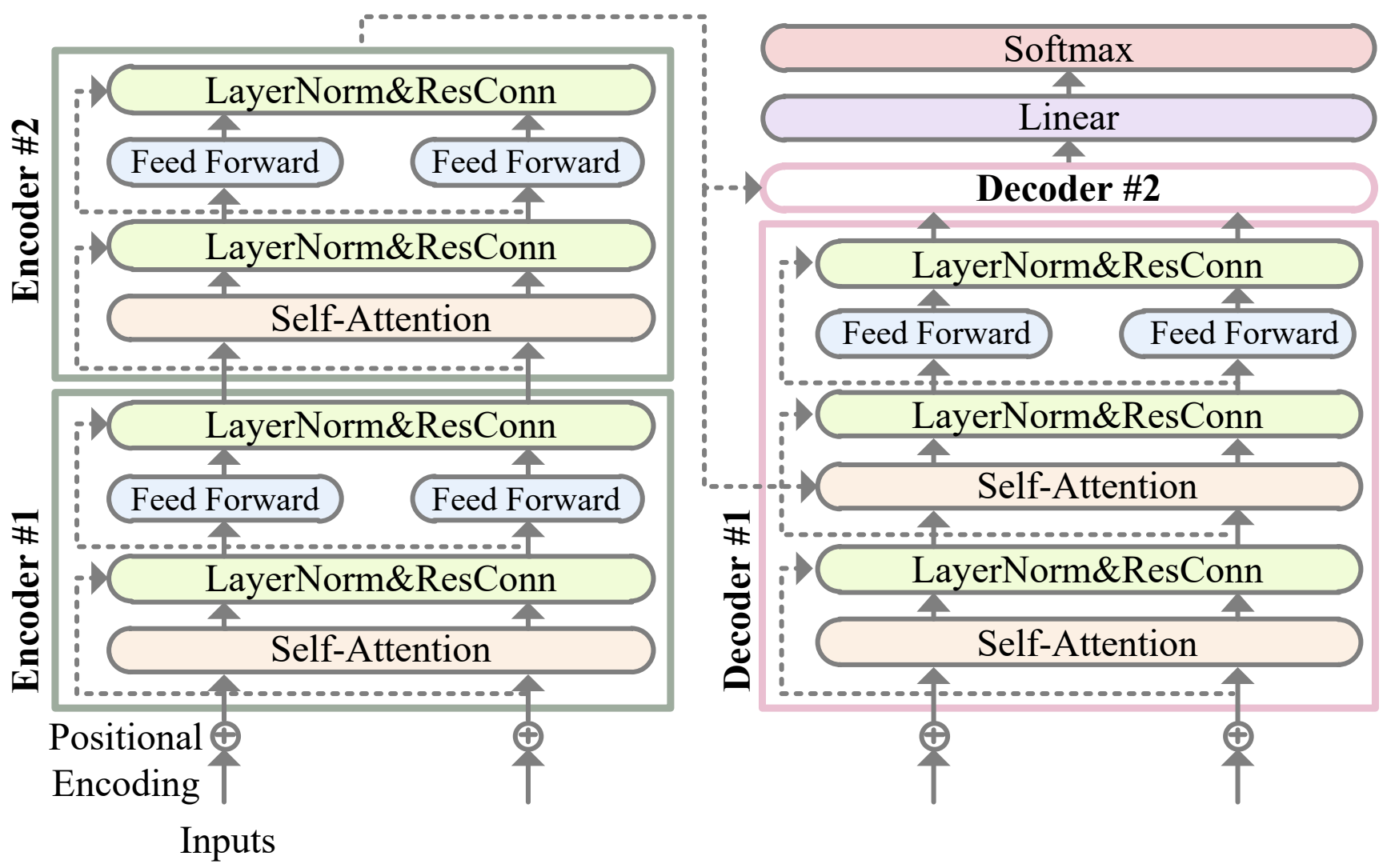

Fig. 3. The text-to-text transfer transformer (T5) architecture.

### *3.3 Model Structure*

To ensure high performance, LLMs are constructed with the text to text transfer transformer (T5) architecture, which has demonstrated highly effective in capturing temporal dependencies and complicated patterns from massive data [24].

As shown in Fig. 3, the T5 architecture consists of an encoder and a decoder. In particular, the encoder transforms the input into a context-aware representation that captures the temporal dependencies of the input, while the decoder uses this representation to generate the output sequence step-by-step. The key components of the T5 architecture include positional coding, self-attention, feed-forward neural network, residual connections, and layer normalization. The following is a brief introduction to each component and its formulation.

***1) Positional Encoding***

In order to assist the model in accurately capturing the sequential information contained within the input features, the function of positional encoding is to provide positional context to each feature through the use of sine-cosine positional encoding [12],[24]. The mathematical formulation of the positional encoding is as follows:

$$\begin{cases} \mathrm{PE}(pos, 2i) = \sin\left(\dfrac{pos}{10000^{2i/d_{\text{model}}}}\right) \\ \mathrm{PE}(pos, 2i+1) = \cos\left(\dfrac{pos}{10000^{2i/d_{\text{model}}}}\right) \end{cases} \tag{6}$$

where $d_{model}$ is the total dimension of the input embedding space; $i$ is the dimension of the embedding; and $pos$ is the position of the token in the sequence.

In Eq. (6), the sine function controls even indices (2i), while the cosine function adjusts odd indices (2i+1). As the position $pos$ increases, the values of PE also vary. This allows the model to differentiate between features at various positions, enabling a more accurate understanding of the input sequence.

***2) Self-Attention***

In the T5 architecture, self-attention serves as the core mechanism, which allows the model to capture diverse relationships and interactions between tokens in the input sequence. The mathematical formulation of the self-attention is as follows:

$$\text{Attention}(Q,K,V)=\text{softmax}\left(\frac{QK^T}{\sqrt{d_q}}\right)V \tag{7}$$

$$\begin{cases} Q=W^Q X \\ K=W^K X \\ V=W^V X \end{cases} \tag{8}$$

where $d_q$ is the dimensionality of the query matrix; $V$ is the value matrix; $K$ is the key matrix; $Q$ is the query matrix; $W^V$, $W^K$, and $W^Q$ are weight matrices used for the linear transformations; and $X$ is the input matrix.

***3) Layer Normalization and Residual Connections***

To improve training stability and convergence, the layer normalization (LayerNorm) is employed to normalize activations within a layer. This process ensures that the inputs to each layer maintain a consistent mean and variance, regardless of batch size, allowing for more effective learning across different training conditions. The mathematical formulation of the layer normalization is as follows:

$$\text{LayerNorm}(x)=\frac{x-\mu}{\sigma}\times\gamma \tag{9}$$

where $\mu$ is the mean of the input; $\sigma$ is the standard deviation of the input; and $\gamma$ is a learnable scale parameter.

To mitigate the vanishing gradient problem and speed up convergence during training, residual connection (ResConn) is used to allow gradients to flow directly through the layers. The mathematical formulation of the residual connection is as follows:

$$\text{ResConn}(x)=F(x)+x \tag{10}$$

where $F(x)$ is the output of the layer $F$.

***4) Feed-Forward Neural Network***

To enable nonlinear transformations and enhance the model's ability to learn complex representations, the feed-forward neural network processes each token independently by using two dense layers to transform the output of the attention mechanism. The mathematical formulation of the feed-forward neural network is as follows:

$$\text{Dense}(x)=\text{ReLU}\left(\left(\text{ReLU}\left(xW_1+B_1\right)\right)W_2+B_2\right) \tag{11}$$

where $W_1$ and $W_2$ are the weight matrices of the 1st and 2nd dense layers, respectively; $B_1$ and $B_2$ are the bias vector of the 1st and 2nd dense layers, respectively; and ReLU is the linear rectification function.

### *3.4 Model Training*

The Chronos model conducts time series forecasting by framing it as a classification problem, where it learns to estimate a distribution over tokens. By reducing the categorical cross-entropy loss between the forecast and actual token distributions, the model is able to produce accurate probabilistic forecasts. Therefore, the loss function of the Chronos model is the categorical cross-entropy loss, which can be formulated as follows:

$$\text{Loss}=-\sum_{t=1}^{T}\sum_{i=1}^{N}y_{t,i}\log\left(\hat{y}_{t,i}\right) \tag{12}$$

where $N$ is the number of possible tokens in the quantized vocabulary (i.e., the number of the bin in Eq.(2)); $T$ is the number of time steps; $y_{t,i}$ is true probability of token $i$ at time step $t$; and $\hat{y}_{t,i}$ is true probability of token $i$ at time step $t$.

The Chronos model is trained on 28 time series datasets, which are gathered from several publicly accessible sources, including Kaggle [26], the M-competitions [27], and the Monash Time Series Repository [28]. These datasets cover a variety of domains such as finance, healthcare, weather, web, retail, energy, and transportation, with time horizons ranging from every 5 minutes to yearly data points. Overall, these datasets contain approximately 890000 univariate time series, with a total of around 84 billion individual observations (tokens).

The model training relies on the Amazon Web Services (AWS) EC2 instance equipped with 8 A100 (40GB) GPUs. The programming language is PyTorch. During training, the optimizer is the AdamW algorithm with a weight decay of 0.01. The learning rate is reduced linearly from an initial value of 0.001 to 0 [24]. All other hyper-parameters in the model are kept at their default values as defined in the Transformers library.

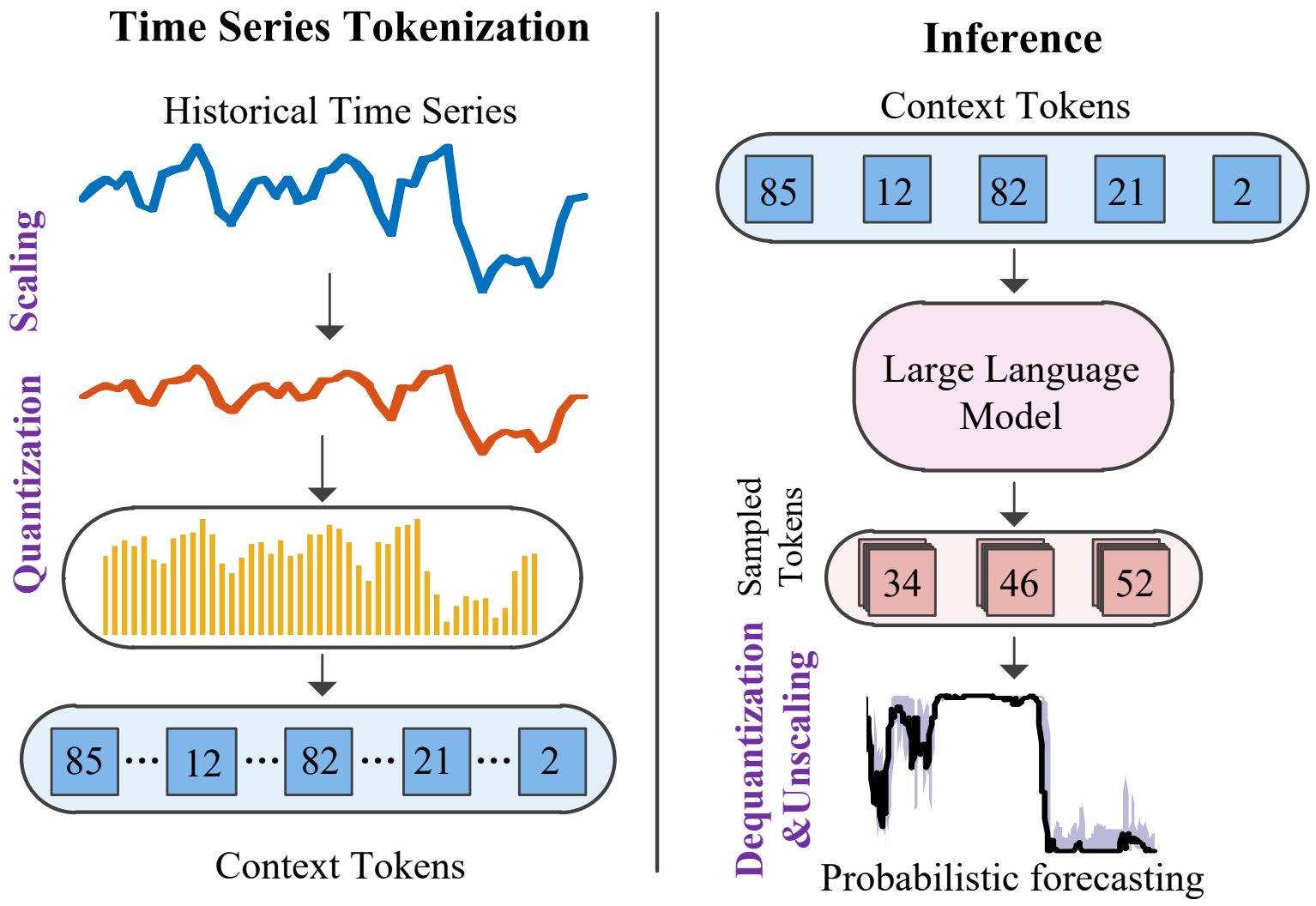

Fig. 4. The main processes of the Chronos model in probabilistic forecasting.

In summary, Fig. 4 visualizes the main processes of the Chronos model in probabilistic forecasting. Notably, deterministic forecasting is also discussed in this paper, represented as a special case of probabilistic forecasting at the 0.5 quantile level.

First, the input time series is normalized and quantized into a sequence of tokens. These tokens are then processed by an LLM framework (i.e., T5 architecture). The parameters of the T5 architecture are optimized by using the cross-entropy loss. After the model training is completed, the historical load data is fed into the Chronos model. The model automatically discretizes the continuous historical load data, which serves as the input for the LLM. The discrete values output by the LLM are then mapped back to continuous values, serving as the final forecast values of loads.

### *3.5 Model Fine-Tuning*

While the Chronos model is pre-trained on large-scale time series datasets, it can be further adapted to specific load forecasting tasks with limited data using fine-tuning techniques. In particular, the widely used low-rank adaptation (LoRA) is applied to reduce the number of trainable parameters while maintaining the original model knowledge. LoRA injects trainable low-rank matrices into the weight matrices of the pre-trained model, allowing efficient adaptation to new tasks with only a small number of additional parameters.

Given a pre-trained weight matrix $W_0 \in \mathbb{R}^{d\times k}$ in a T5 layer (e.g., in self-attention or feed-forward layers), LoRA introduces two low-rank matrices $A \in \mathbb{R}^{d\times r}$ and $B \in \mathbb{R}^{r\times k}$, where min($d$,$k$). During fine-tuning, only $A$ and $B$ are trained, while $W_0$ remains frozen. The adapted weight is formulated as:

$$W = W_0 + \alpha AB \tag{13}$$

where α is a scaling factor that controls the contribution of the low-rank update.

For input token embeddings $X \in \mathbb{R}^{T\times d}$ (where T is the sequence length), the linear transformation with LoRA can be written as:

$$Y = XW_0^\top + \alpha XB^\top A^\top \tag{14}$$

This allows the model to adjust its output according to the new load data while preserving the pre-trained knowledge from large datasets.

Given a small load dataset with $T$ time steps and quantized tokens, the fine-tuning is performed by minimizing the categorical cross-entropy loss over the LoRA parameters:

$$\mathcal{L}_{\text{LoRA}} = -\sum_{t=1}^{T}\sum_{i=1}^{N} y_{t,i} \log \hat{y}_{t,i}\left(W_0 + \alpha AB\right) \tag{15}$$

where $\hat{y}_{t,i}$ is the predicted probability of token $i$ at time step $t$; N is the number of discrete tokens (bins), and $y_{t,i}$ is the true token distribution.

During fine-tuning, only the low-rank matrices $A$ and $B$ are updated with the limited target load dataset. This significantly reduces memory and computation requirements while enabling the model to adapt to new domains or buildings. After training, the adapted model is used in the same forecasting pipeline: input load sequences are scaled and quantized, processed by the LoRA-enhanced T5, and the output tokens are mapped back to continuous load values for final predictions.

### *3.6 Forecasting with Covariates*

The Chronos model described above primarily focuses on univariate time series forecasting, where future load values are predicted based on historical load observations. However, in practical load forecasting applications, electricity demand is not only driven by its historical patterns but also influenced by various external factors. To better capture such influences, the Chronos framework can be naturally extended to incorporate covariates, enabling forecasting with exogenous variables within the same token-based LLM architecture.

Let $\{y_t\}_{t=1}^{T}$ denote the historical load time series, and let $\{\mathbf{z}_t\}_{t=1}^{T}, \mathbf{z}_t \in \mathbb{R}^C$ represent the corresponding covariate series. In this extended setting, the forecasting task aims to learn the conditional distribution of future load values given both historical load data and covariate information, which can be expressed as:

$$p\left(y_{t+1:t+H} \mid y_{1:t}, \mathbf{z}_{1:t+H}\right) \tag{16}$$

where $H$ is the forecasting horizon. This formulation allows the model to condition its predictions on known or forecasted exogenous variables.

To integrate covariates into the Chronos framework, the covariate time series are processed using the same scaling and quantization procedures as the target load series, transforming continuous exogenous variables into discrete tokens. These covariate tokens are then combined with the load tokens to form a unified input sequence for the T5-based architecture. Through the self-attention mechanism, the model jointly captures temporal dependencies in the load series and the interactions between load and covariates, without requiring explicit feature engineering or predefined dependency structures.

In the context of load forecasting, incorporating covariates is particularly important due to the strong dependence of electricity demand on meteorological conditions. Variables such as temperature, humidity, wind speed, and solar irradiance play a critical role in shaping daily and seasonal load patterns, as well as in driving abrupt demand changes during extreme weather events. By integrating these weather-related covariates into the Chronos model, the forecasting framework becomes more expressive and robust, leading to improved probabilistic load forecasts under diverse operating conditions.

### *3.7 Implementation Details*

All load time series are preprocessed following a unified pipeline. Missing values are handled by using linear interpolation. If a missing segment exceeds five consecutive time steps, the corresponding sample is excluded from training. Extreme outliers are detected using the interquartile range (IQR) rule and are clipped to the boundary values to avoid distortion of the scaling process. After cleaning, mean scaling is applied independently to each time series, followed by percentile-based quantization with 100 bins, as described in Section 3.1. The same preprocessing pipeline is consistently used during training, fine-tuning, and inference.

Chronos is pre-trained as described in Section 3.4. For task-specific adaptation, fine-tuning is conducted using LoRA, where only the low-rank parameters are updated while the backbone T5 model remains frozen. The batch size is set to 64, and the AdamW optimizer is used. The learning rate for LoRA parameters is $1 \times 10^{-4}$. The maximum number of fine-tuning epochs is 50, with early stopping applied using a patience of 5 epochs based on validation loss. To ensure reproducibility, all experiments are conducted with a fixed random seed of 42.

Forecasting is performed using a rolling-window strategy. For each prediction time step, the most recent lookback window is used as input. Multi-step forecasting is conducted in a direct manner, where the full forecasting horizon is predicted in a single forward pass, rather than recursively feeding back intermediate predictions. The stride between consecutive forecasting windows is set to one time step. A warm-up period equal to the lookback window length is required before generating the first forecast.

For probabilistic forecasting, Chronos directly outputs token-level probability distributions, which are decoded into prediction intervals and quantiles. Deterministic forecasts are obtained as a special case by selecting the median (0.5 quantile) of the predictive distribution. All evaluation metrics reported in Section 4 are computed using the same inference procedure to ensure a fair and consistent comparison.

## 4 Case Study I

### 4.1 Simulation Settings

***1) Dataset Description***

Simulations are conducted on a real-world load dataset from the University of Texas at Austin [29]. This dataset has not been used to parameterize the LLM, ensuring the zero-shot load forecasting. It includes hourly load data for 16 campus buildings from August 1, 2011, to July 31, 2012.

For all experiments, the data split is strictly chronological, and no shuffling is applied at any stage. The training, validation, and test sets are separated according to their original time order to prevent any leakage of future information into the training process. The forecasting horizons considered in this paper range from 1 h to 48 h, including 1, 6, 12, 24, and 48 hours, as detailed in Sections 4.2–4.7. For each forecast horizon, a rolling-origin evaluation strategy is adopted on the test period. To evaluate the model performance in different scenarios, two scenarios are designed: a data-rich scenario and a data-scarce scenario, as shown in Table I.

TABLE I
DESCRIPTION OF TWO DATA SCENARIOS

| Scenario | Training set and validation set | Test set |
|---|---|---|
| Data-rich scenario | Aug. 1, 2011, to Jan. 31, 2012 (6 months) | Feb. 1, 2012, to Jul. 31, 2012 (6 months) |
| Data-scarce scenario | Jan. 17, 2012, to Jan. 31, 2012 (15 days) | Feb. 1, 2012, to Jul. 31, 2012 (6 months) |

In the data-rich scenario, the training and validation data span from August 1, 2011, to January 31, 2012 (6 months, with 80% used for training and 20% for validation), while the test data range from February 1, 2012, to July 31, 2012 (6 months). This ensures sufficient data for training machine learning models, excluding LLMs, which are already pre-trained and are employed here for zero-shot load forecasting. In other words, the load data from the University of Texas at Austin will be not used to update the parameters of LLMs.

In the data-scarce scenario (may be due to factors like newly developed areas or privacy concerns), the training and validation data are drawn from January 17, 2012, to January 31, 2012 (15 days, with 80% for training and 20% for validation). The test data remain the same, covering February 1, 2012, to July 31, 2012 (6 months). Similarly, the load data here is only used to train statistical models as well as deep learning models, excluding LLMs.

To clarify the training regimes and ensure a fair comparison, it is emphasized that different models follow different data usage strategies consistent with their design principles. Specifically, statistical models and deep learning models (e.g., TFT, DeepAR, and PatchTST) are trained exclusively on the target dataset using the corresponding training set defined in Table I, while the test set is strictly held out and never used for model training or hyperparameter tuning. In contrast, pre-trained LLM-based models, including Chronos and TimeLLM, are **not trained on the target load dataset** in the zero-shot setting. Instead, they rely solely on knowledge acquired during large-scale pre-training on external time series corpora. In the data-scarce scenario, the limited target data are only used to train statistical and deep learning baselines, while Chronos is evaluated in a zero-shot or few-shot manner as explicitly stated. Throughout all experiments, no future information from the test period is accessed during training or validation, ensuring the absence of data leakage.

***2) Baseline Models***

To perform the comparative analysis, baseline models include popular statistical models (e.g., AutoETS [3], SNM [4], CSBA [5], NPTS [6], and AutoARIMA [7]), advanced neural networks (e.g., TFT [12], DeepAR [11], and PatchTST [13]), and a pre-trained LLM (e.g., TimeLLM [22]).

- AutoETS [3]: This is a statistical model. For deterministic forecasting, AutoETS selects the optimal combination of error, trend and seasonality components to decompose time series data and provide a single future value. For probabilistic forecasting, it utilizes the statistical properties of these components to determine a forecast distribution that provides the probability interval within which future values are likely to fall.
- SNM [4]: This is a statistical model. For deterministic forecasting, it assumes that the forecast value is equal to the last historical value from the same season. For probabilistic forecasting, quantiles are derived under the assumption that the residuals follow a zero-mean normal distribution, with the scale being determined based on the empirical distribution of the residuals.
- CSBA [5]: This is a widely used statistical model for forecasting intermittent demand. It decomposes the demand process into two components: the demand magnitude and the interval between non-zero demands, which are modeled separately. The final forecast is achieved by combining these components. The syntetos and boylan approximate (SBA) variant introduces a bias correction to the original Croston model, which improves the accuracy of the probabilistic forecasting.
- NPTS [6]: This is a flexible statistical model. For deterministic forecasting, NPTS forecasts future values by finding similar historical data that resemble the current observations. For probabilistic forecasting, NPTS generates quantiles by sampling from historical data rather than assuming a fixed distribution, making the model highly flexible and adaptable to different data patterns.
- AutoARIMA [7]: This is a statistical model. The automatically tuned ARIMA model selects the optimal parameters using an information criterion. This automated process improves forecast accuracy and simplifies model selection.
- TFT [12]: This is a transformer-based neural network, which integrates both recurrent neural networks and attention mechanisms specifically designed for time-series forecasting. Its architecture automatically captures temporal patterns and relationships in the data, enhancing forecast accuracy. The probabilistic forecasting is conducted by the pinball loss function [2].
- DeepAR [11]: This is a recurrent neural network-based autoregressive forecasting model. For deterministic forecasting, the model uses the long short-term memory to model temporal relationships. For probabilistic forecasting, it generates the full probability distribution by sampling from the output of the long short-term memory, assuming that future values follow a specified parametric distribution.
- PatchTST [13]: This is a transformer-based deep learning model. For deterministic forecasting, it divides the time series into patches and uses self-attention mechanisms to capture short-term dependencies across the time series. For probabilistic forecasting, it is achieved by using the pinball loss function.
- TimeLLM [24]: This is a reprogramming framework to repurpose LLMs for time series forecasting. The principle is to convert time series data into text and use the capabilities of LLMs for sequence modeling to capture complex temporal dependencies and patterns.
- Chronos: The details of model training have been introduced in Section 3.4.

To ensure a fair comparison, the parameters of each model are determined by Bayesian optimization. To ensure a fair comparison with TimeLLM, we have optimize the recommended prompt settings consistently across all forecast horizons and datasets. For SNM, the seasonal period is automatically inferred from the frequency of the training data. For NPTS, the kernel used by the model is an exponential kernel. The seasonal variant of the model is used. The number of time features used by the seasonal model is 1. For AutoARIMA, the maximum number of autoregressive terms and moving average orders is 5 each. It allows up to 2 seasonal autoregressive terms and 2 seasonal moving average orders, and a maximum of 2 for first differencing. For TFT, the batch size is 64, and the max epoch is 100. The dropout rate is 0.1, and the number of attention heads in self-attention is 4. The size of the feature embedding is 32, and the size of the LSTM and transformer hidden states is 32. For DeepAR, The number of recurrent neural network (RNN) cells for each layer is 40, and the number of RNN layers is 2. Other training parameters are consistent with TFT. For PatchTST, the length of the patch is 16, and the stride of the patch is 8. The number of attention heads in the transformer encoder is 4, and the size of hidden layers in the transformer encoder is 32. The number of layers in the transformer encoder is 2. Other training parameters are consistent with TFT. For TimeLLM, the length of the patch is 16, and the stride of the patch is 8. The top token to consider is 5, and the hidden dimension of LLM is 768. The number of heads in the attention layer is 8.

The choice of the lookback window needs to vary according to the forecast horizon. To determine the optimal lookback window for each forecast horizon, Bayesian optimization is employed. To ensure a fair comparison, the lookback window used by each model remains consistent.

To further ensure fairness and comparability, all models are provided with identical input information, including the same target load series, identical covariates (when applicable), and the same forecast horizons. The lookback window for each forecast horizon is determined via Bayesian optimization and then fixed consistently across all models. No model is allowed access to future signals beyond the forecasting origin.

Hyperparameters of all trainable models are optimized only on the training and validation sets, and the test set is strictly reserved for final performance evaluation. For TimeLLM, we optimize prompt-related settings consistently across all datasets and horizons, without using any test data. Therefore, although the training paradigms differ (fully trained, zero-shot, or few-shot), the comparison protocol is fair and meaningful, reflecting realistic deployment scenarios under both data-rich and data-scarce conditions.

The algorithms are implemented in Spyder. The system features an Intel(R) Core(TM) i5-10210U CPU operating at a base frequency of 1.60 GHz and has 8 GB of RAM.

***3) Evaluation metrics***

For deterministic load forecasting, the widely used root mean squared error (RMSE), mean absolute error (MAE), and mean absolute percentage error (MAPE) are employed to evaluate the model performance. The smaller values of RMSE, MAE, and MAPE indicate better model performance in terms of forecast accuracy. Their mathematical formulations are as follows [1],[2],[30]:

$$\mathrm{RMSE} = \sqrt{\frac{1}{M}\sum_{i=1}^{M}(\hat{x}_i - x_i)^2} \tag{17}$$

$$\mathrm{MAE} = \frac{1}{M}\sum_{i=1}^{M}\left|\hat{x}_i - x_i\right| \tag{18}$$

$$\mathrm{MAPE} = \frac{100\%}{M}\sum_{i=1}^{M}\left|\frac{\hat{x}_i - x_i}{x_i}\right| \tag{19}$$

where $\hat{x}_i$ is the actual value; $x_i$ is the forecast value; and $M$ is the number of data points in the test set.

For probabilistic load forecasting, the widely used prediction interval coverage probability (PICP), prediction interval average width (PIAW), average coverage error (ACE), Winkler score (WS), quantile score (QS), and continuous ranked probability score (CRPS) are employed to evaluate the model performance. The smaller values of QS, WS, ACE, and CRPS indicate better probabilistic forecast performance, as they measure the accuracy and sharpness of the forecast distribution. Meanwhile, higher PICP and lower PINAW indicate a well-calibrated and narrower prediction interval (PI), balancing coverage and interval width. Their mathematical formulations are as follows [1],[2],[10]:

$$\mathrm{PICP}_{\alpha} = \frac{1}{M}\sum_{i=1}^{M}c_i, \quad c_i = \begin{cases} 1, x_i \in [L_{\alpha,i}, U_{\alpha,i}] \\ 0, x_i \notin [L_{\alpha,i}, U_{\alpha,i}] \end{cases} \tag{20}$$

$$\mathrm{PIAW}_{\alpha} = \frac{1}{M}\sum_{i=1}^{M}\left(U_i^{\alpha} - L_i^{\alpha}\right) \tag{21}$$

$$\mathrm{ACE}_{\alpha} = \mathrm{PICP}_{\alpha} - \mathrm{PINC}_{\alpha} \tag{22}$$

$$\mathrm{WS}_{\alpha} = \frac{1}{M}\sum_{i=1}^{M} \mathrm{WS}_{\alpha,i} \tag{23}$$

$$\mathrm{WS}_{\alpha,i} = \begin{cases} U_{\alpha,i} - L_{\alpha,i}, & L_{\alpha,i} < x_i < U_{\alpha,i} \\ U_{\alpha,i} - L_{\alpha,i} + 2(x_i - U_{\alpha,i})/\alpha, & x_i > U_{\alpha,i} \\ U_{\alpha,i} - L_{\alpha,i} + 2(L_{\alpha,i} - x_i)/\alpha, & x_i < L_{\alpha,i} \end{cases} \tag{24}$$

$$\mathrm{QS} = \frac{1}{N_{\alpha}}\sum_{i=1}^{N_{\alpha}}\left(\frac{1}{M}\sum_{i=1}^{M} \text{Pinball loss}\left(x_i, \hat{x}_{\alpha,i}\right)\right) \tag{25}$$

$$\mathrm{CRPS}(F, x) = \int_{-\infty}^{+\infty} \left(F(z) - I(z \geq x)\right)^2 dz \tag{26}$$

where $\alpha$ is the quantile (e.g., 0.1, 0.2,…,0.9); $L_{\alpha,i}$ is the lower boundary of the PI with prediction interval nominal confidence (PINC) 100(1-$\alpha$)% at the $i^{\text{th}}$ data point; $U_{\alpha,i}$ is the upper boundary of the PI with PINC 100(1-$\alpha$)% at the $i^{\text{th}}$ data point; $N_{\alpha}$ is the number of the quantile; $F(z)$ is the cumulative distribution function of the forecasts; $z$ is a variable representing the range of possible forecast values, used to calculate the error; $I(z \geqslant x)$ is an indicator function, which equals 1 when $z \geqslant x$, and 0 otherwise; and $\hat{x}_{\alpha,i}$ is the $i^{\text{th}}$ forecast value of the $\alpha$-quantile.

### *4.2 Performance Comparison in Data-Rich Scenarios*

In this section, the performance of the Chronos model is compared with baseline models in a data-rich scenario. The forecast horizons are set at 1, 6, 12, 24, and 48 hours. Each model is executed 30 times, and the average test set error is calculated, as presented in Table II.

TABLE II
PERFORMANCE EVALUATION OF DIFFERENT MODELS IN DATA-RICH SCENARIOS

| Model | Forecast horizon is 1 h | | | Forecast horizon is 6 h | | | Forecast horizon is 12 h | | | Forecast horizon is 24 h | | | Forecast horizon is 48 h | | |
|---|---|---|---|---|---|---|---|---|---|---|---|---|---|---|---|
| | RMSE (MW) | MAE (MW) | MAPE (%) | RMSE (MW) | MAE (MW) | MAPE (%) | RMSE (MW) | MAE (MW) | MAPE (%) | RMSE (MW) | MAE (MW) | MAPE (%) | RMSE (MW) | MAE (MW) | MAPE (%) |
| SNM | 3.79 | 2.83 | 6.95% | 3.79 | 2.83 | 6.95% | 3.79 | 2.83 | 6.95% | 3.79 | 2.83 | 6.95% | 4.34 | 3.31 | 8.09% |
| CSBA | 4.26 | 3.41 | 7.96% | 4.90 | 3.90 | 9.16% | 5.33 | 4.44 | 10.53% | 4.60 | 3.73 | 8.96% | 5.05 | 4.07 | 9.76% |
| NPTS | 5.01 | 4.05 | 9.44% | 5.03 | 4.06 | 9.46% | 5.02 | 4.05 | 9.43% | 5.01 | 4.06 | 9.46% | 5.05 | 4.09 | 9.53% |
| AutoETS | 1.18 | 0.89 | 2.17% | 3.69 | 2.82 | 6.88% | 3.52 | 2.81 | 7.05% | 4.32 | 3.46 | 8.36% | 4.80 | 3.84 | 9.32% |
| AutoARIMA | 1.02 | 0.77 | 1.90% | 2.94 | 2.14 | 5.28% | 3.55 | 2.66 | 6.56% | 4.28 | 3.19 | 7.81% | 4.41 | 3.42 | 8.39% |
| TFT | 0.85 | 0.61 | 1.48% | 1.82 | 1.31 | 3.17% | 2.20 | 1.62 | 3.97% | 3.17 | 2.33 | 5.83% | 3.87 | 2.88 | 7.00% |
| DeepAR | 1.11 | 0.83 | 2.00% | 2.69 | 2.01 | 4.69% | 2.18 | 1.63 | 4.00% | 3.79 | 3.00 | 6.91% | 10.40 | 8.44 | 19.07% |
| PatchTST | 0.87 | 0.63 | 1.53% | 2.18 | 1.53 | 3.70% | 2.94 | 2.25 | 5.63% | 3.23 | 2.41 | 5.95% | 3.56 | 2.74 | 6.71% |
| TimeLLM | 4.56 | 3.67 | 8.75% | 6.19 | 4.95 | 12.47% | 7.08 | 5.54 | 14.05% | 10.64 | 8.39 | 20.79% | 8.07 | 6.43 | 15.98% |
| Chronos | **0.79** | **0.56** | **1.36%** | **1.59** | **1.10** | **2.68%** | **2.08** | **1.49** | **3.66%** | **2.48** | **1.82** | **4.42%** | **2.93** | **2.14** | **5.26%** |

#### ***1) Performance Comparison***

The results show that the Chronos model outperforms the baseline models at different forecast horizons (e.g., 1 hour, 6 hours, 12 hours, 24 hours, and 48 hours). For example, at the forecast horizon of 1 hour, the Chronos model achieves RMSE, MAE, and MAPE values of 0.79 MW, 0.56 MW, and 1.36%, respectively. Compared to the baseline models (i.e., SNM, CSBA, NPTS, AutoETS, AutoARIMA, TFT, DeepAR, PatchTST, and TimeLLM), Chronos reduces the RMSE, MAE, and MAPE by approximately 7.34%-84.30%, 8.96%-86.21%, and 7.87%-85.60% respectively.

For longer forecast horizons (e.g. 24 hours and 48 hours), the Chronos model continues to maintain low forecast errors. At the forecast horizon of 48 hours, Chronos achieves RMSE, MAE, and MAPE values of 2.93 MW, 2.14 MW, and 5.26%, respectively. Compared to these baseline methods, the Chronos model reduces the RMSE, MAE, and MAPE by approximately 17.72%-71.82%, 21.81%-74.59%, and 21.57%-72.42% respectively. This demonstrates that Chronos remains both robust and accurate for load forecasting with longer forecast horizons (e.g. 24 hours and 48 hours).

Overall, the Chronos model shows superior performance in load forecasting, particularly in short-term horizons, where it significantly reduces forecast errors while offering enhanced robustness and generalization capabilities compared to existing models.

***2) Zero-Shot Performance of Chronos Model***

Chronos is a pre-trained LLM with parameters determined from extensive time series data from multiple sources, without any fine-tuning for this specific load dataset from the University of Texas at Austin. Thus, it functions as a zero-shot load forecasting model. In contrast, most baseline models (e.g., statistical and deep learning models) have been specifically trained and optimized using this dataset. Despite not being tailored or fine-tuned to this dataset, the Chronos model outperforms these customized models, demonstrating its strong generalization capabilities to unseen data.

While TimeLLM is also a pre-trained LLM (i.e., a zero-shot load forecasting model), its performance falls significantly short of the Chronos model. This discrepancy is due to fundamental differences in how the two models are designed and trained. Specifically, TimeLLM is trained on textual data, and it handles time-series data by first converting it into text, feeding this text into the LLM, and then converting the output back into time-series data. In contrast, the Chronos model uses the LLM framework but is directly pre-trained on massive time series data rather than text data. This direct training on time series data gives Chronos a significant advantage in handling time series analysis tasks, resulting in superior performance load forecasting compared to other LLMs, such as TimeLLM.

***3) Visual Analysis***

For the visual analysis, 3 days of load data are randomly selected from the dataset for different forecast horizons (e.g., 1 hour, 6 hours, 24 hours, and 48 hours). Then, the actual load values are compared with the forecast values from various models, including the Chronos model and baseline models, as shown in Fig. 5.

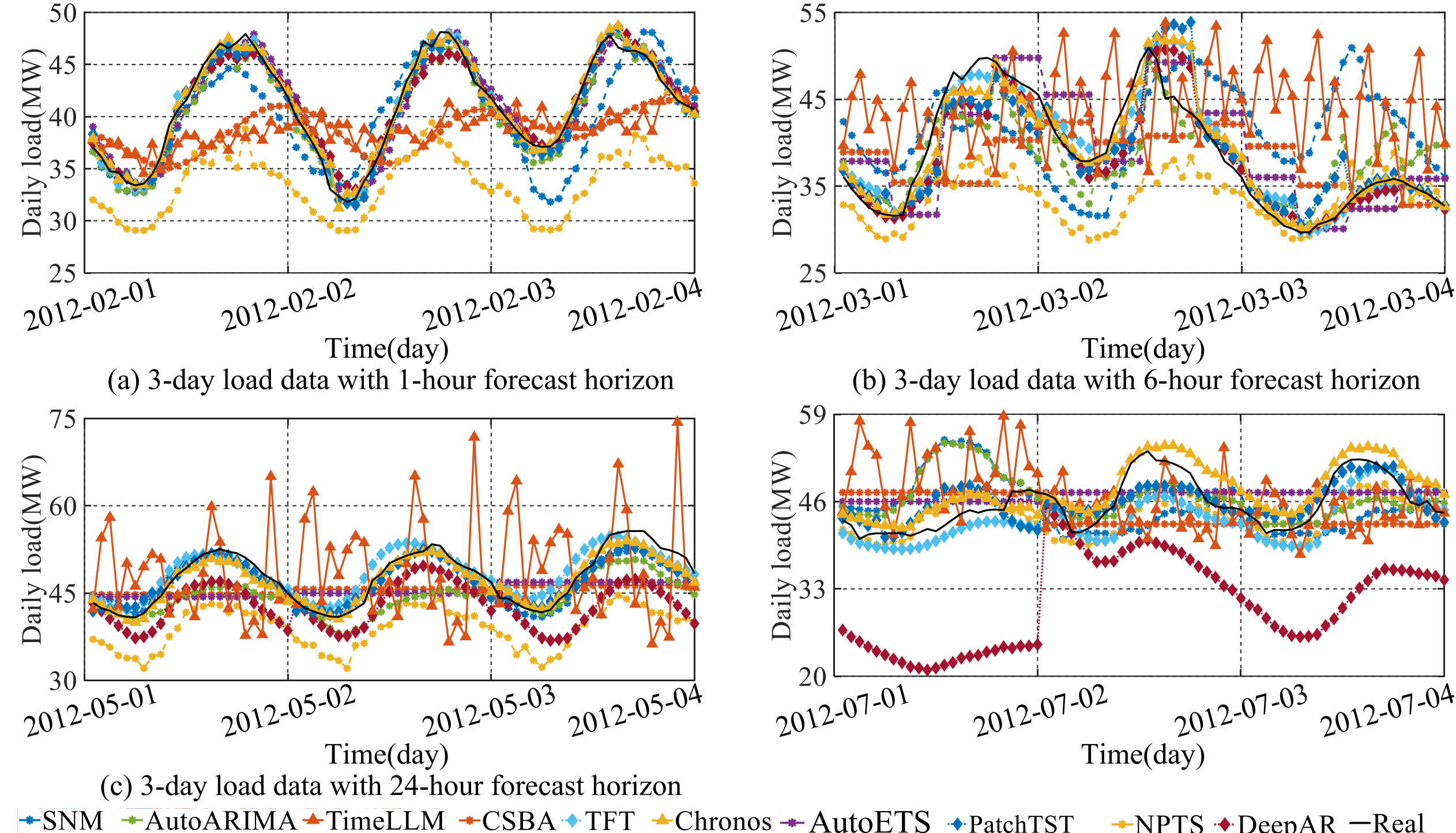

(a) 3-day load data with 1-hour forecast horizon

(b) 3-day load data with 6-hour forecast horizon

(c) 3-day load data with 24-hour forecast horizon

Fig. 5. Visual comparison of load forecasting models across different forecast horizons.

For very short forecast horizons (e.g. 1 hour), the forecast values of most models are close to the actual values, indicating high accuracy. However, as the forecast horizon increases, especially for the 24-hour and 48-hour scenarios, the forecast errors of baseline models increase significantly, showing their limitations in long forecast horizons.

In contrast, the Chronos model consistently performs well across both short and long forecast horizons. Whether for 1 hour or 48 hours, its forecast values remain closely aligned with the actual values, and it also significantly outperforms the baseline models. This highlights the superior robustness and accuracy of the Chronos model across various forecast horizons.

### *4.3 Performance Comparison in Data-Scarce Scenarios*

Sometimes, enough data is not available to train forecasting models due to various factors like newly developed areas or privacy concerns. In this section, the Chronos model is compared with baseline models in a data-scarce scenario. The forecast horizons are set at 1, 6, 12, 24, and 48 hours. Each model is executed 30 times, and the average test set error is calculated, as presented in Table III.

TABLE III
PERFORMANCE EVALUATION OF DIFFERENT MODELS IN DATA-SCARCE SCENARIOS

| Model | Forecast horizon is 1 h | | | Forecast horizon is 6 h | | | Forecast horizon is 12 h | | | Forecast horizon is 24 h | | | Forecast horizon is 48 h | | |
|---|---|---|---|---|---|---|---|---|---|---|---|---|---|---|---|
| | RMSE (MW) | MAE (MW) | MAPE (%) | RMSE (MW) | MAE (MW) | MAPE (%) | RMSE (MW) | MAE (MW) | MAPE (%) | RMSE (MW) | MAE (MW) | MAPE (%) | RMSE (MW) | MAE (MW) | MAPE (%) |
| SNM | 3.79 | 2.83 | 6.95% | 3.79 | 2.83 | 6.95% | 3.79 | 2.83 | 6.95% | 3.79 | 2.83 | 6.95% | 4.34 | 3.31 | 8.09% |
| CSBA | 4.26 | 3.41 | 7.96% | 4.90 | 3.90 | 9.16% | 5.33 | 4.44 | 10.53% | 4.60 | 3.73 | 8.96% | 5.05 | 4.07 | 9.76% |
| NPTS | 3.96 | 3.17 | 7.77% | 3.96 | 3.17 | 7.77% | 3.96 | 3.18 | 7.79% | 3.95 | 3.17 | 7.78% | 4.09 | 3.28 | 8.03% |
| AutoETS | 1.18 | 0.89 | 2.17% | 3.69 | 2.82 | 6.88% | 3.52 | 2.81 | 7.05% | 4.32 | 3.46 | 8.36% | 4.80 | 3.84 | 9.32% |
| AutoARIMA | 0.92 | 0.69 | 1.69% | 2.64 | 1.93 | 4.74% | 3.21 | 2.42 | 5.96% | 3.69 | 2.85 | 6.96% | 4.07 | 3.18 | 7.82% |
| TFT | 0.83 | 0.59 | 1.44% | 2.42 | 1.77 | 4.30% | 3.52 | 2.67 | 6.66% | 4.37 | 3.40 | 8.61% | 5.13 | 3.97 | 9.88% |
| DeepAR | 1.76 | 1.37 | 3.34% | 3.21 | 2.54 | 6.26% | 3.44 | 2.74 | 6.63% | 4.82 | 3.92 | 9.95% | 7.41 | 6.19 | 15.53% |
| PatchTST | 1.09 | 0.84 | 2.04% | 2.38 | 1.84 | 4.45% | 2.92 | 2.19 | 5.33% | 3.88 | 3.06 | 7.59% | 5.79 | 4.65 | 10.99% |
| TimeLLM | 4.51 | 3.64 | 8.67% | 5.66 | 4.56 | 11.37% | 7.18 | 5.71 | 14.45% | 8.70 | 7.04 | 17.59% | 6.87 | 5.55 | 13.87% |
| Chronos | **0.79** | **0.56** | **1.36%** | **1.59** | **1.10** | **2.68%** | **2.08** | **1.49** | **3.66%** | **2.48** | **1.82** | **4.42%** | **2.93** | **2.14** | **5.26%** |

***1) Model Performance Analysis***

For statistical models (e.g., SNM, CSBA, NPTS, AutoETS, and AutoARIMA), despite the scarcity of data, their forecast accuracy does not vary significantly. This is because statistical models typically rely on a few historical values, allowing them to make reasonable inferences even with limited historical data. Thus, they can maintain a certain level of stability and accuracy in data-scarce conditions.

In contrast, deep learning models (TFT, DeepAR, and PatchTST) experience a notable increase in forecast errors in data-scarce scenarios compared to data-rich scenarios, especially in longer forecast horizons (e.g., the forecast horizon is 24 hour or 48 hours). This is because deep learning models require massive training data to learn complex patterns and features. When the data is insufficient, the models are prone to overfitting or may fail to capture the underlying patterns effectively.

Against this backdrop, pre-trained LLMs, such as the Chronos model, demonstrate their advantages in data-scarce scenarios. The Chronos model does not require fine-tuning on load data (i.e., it performs zero-shot load forecasting), allowing it to adapt more effectively in sparse data conditions. This is because the Chronos model has been pre-trained on extensive time series data, and it can maintain high forecast performance even without specific adjustments to a particular dataset.

***2) Performance Comparison***

Furthermore, in data-scarce scenarios, the forecast errors of the Chronos model are significantly lower than those of the baseline models over different forecast horizons.

For example, at the 1-hour forecast horizon, compared to baseline models, Chronos reduces the RMSE, MAE, and MAPE by approximately 4.82%-82.48%, 5.08%-84.62%, and 5.56%-84.31% respectively. At the 48-hour forecast horizon, compared to these baseline methods, the Chronos model reduces the RMSE, MAE, and MAPE by approximately 28.01-60.46%, 32.70%-65.43%, and 32.74%-66.13% respectively.

This further confirms the superiority and flexibility of the Chronos model, making it an effective solution in data-scarce scenarios.

### *4.4 Performance Comparison in Multivariate Forecasting*

In practical load forecasting scenarios, electricity demand is often influenced by exogenous variables, such as NWP data, in addition to historical load observations. Therefore, this section extends the univariate forecasting analysis to forecasting with covariates. The data-scarce scenario from the University of Texas is used as an example, and this section evaluates the impact of incorporating NWP information on forecasting performance. Since SNM, CSBA, NPTS, AutoETS, and AutoARIMA do not support multivariate inputs, the comparison is limited to TFT, DeepAR, PatchTST, TimeLLM, and Chronos. The considered NWP variables include humidity, wind speed, pressure, wind direction, and sky condition. The forecast horizons are set at 1, 6, 12, 24, and 48 hours. Each model is executed 30 times, and the average test set error is calculated, as presented in Table IV.

TABLE IV
PERFORMANCE EVALUATION OF DIFFERENT MODELS IN DATA-SCARCE SCENARIOS WITH NWP DATA

| Model | Forecast horizon is 1 h | | | Forecast horizon is 6 h | | | Forecast horizon is 12 h | | | Forecast horizon is 24 h | | | Forecast horizon is 48 h | | |
|---|---|---|---|---|---|---|---|---|---|---|---|---|---|---|---|
| | RMSE (MW) | MAE (MW) | MAPE (%) | RMSE (MW) | MAE (MW) | MAPE (%) | RMSE (MW) | MAE (MW) | MAPE (%) | RMSE (MW) | MAE (MW) | MAPE (%) | RMSE (MW) | MAE (MW) | MAPE (%) |
| TFT | 0.75 | 0.52 | 1.29% | 2.18 | 1.57 | 3.86% | 3.16 | 2.36 | 5.98% | 3.93 | 3.01 | 7.73% | 4.61 | 3.51 | 8.87% |
| DeepAR | 1.58 | 1.21 | 3.00% | 2.89 | 2.25 | 5.62% | 3.09 | 2.42 | 5.95% | 4.33 | 3.47 | 8.94% | 6.66 | 5.48 | 13.95% |
| PatchTST | 0.98 | 0.74 | 1.83% | 2.14 | 1.63 | 4.00% | 2.63 | 1.94 | 4.79% | 3.49 | 2.71 | 6.82% | 5.21 | 4.11 | 9.87% |
| TimeLLM | 4.05 | 3.22 | 7.79% | 5.09 | 4.03 | 10.21% | 6.45 | 5.05 | 12.98% | 7.82 | 6.23 | 15.80% | 6.18 | 4.91 | 12.46% |
| Chronos | **0.71** | **0.50** | **1.22%** | **1.43** | **0.97** | **2.41%** | **1.87** | **1.32** | **3.29%** | **2.23** | **1.61** | **3.97%** | **2.63** | **1.89** | **4.72%** |

By comparing Table III and Table IV, it can be observed that incorporating NWP data consistently reduces the forecasting errors of all models across different forecast horizons. This indicates that exogenous meteorological information provides valuable complementary information beyond historical load observations, enabling models to better capture load variations.

Moreover, the Chronos model consistently achieves lower forecasting errors than TFT, DeepAR, PatchTST, and TimeLLM across all forecast horizons (e.g., forecast horizons range from 1 h to 48 h). This demonstrates that Chronos is effective in using covariate information while maintaining strong generalization ability. Benefiting from large-scale pre-training on diverse time series data, Chronos exhibits superior adaptability and robustness when incorporating NWP data, making it particularly suitable for practical load forecasting scenarios with limited historical data.

### *4.5 Performance Comparison With and Without Model Fine-Tuning*

In practical load forecasting applications, a small amount of historical data is often available and can be leveraged to further adapt pre-trained models to specific datasets. While Sections 4.2-4.4 evaluate the Chronos model under a zero-shot load forecasting setting, this section investigates the impact of model fine-tuning (i.e., few-shot load forecasting). Specifically, the data-scarce scenario from the University of Texas is used as an example, and this section compares the forecasting performance of Chronos model with and without fine-tuning to assess whether limited task-specific data can further enhance its effectiveness in realistic deployment scenarios. The forecast horizons are set at 1, 6, 12, 24, and 48 hours. The average test errors of Chronos model with and without fine-tuning are calculated, as presented in Table V.

TABLE V
PERFORMANCE EVALUATION OF DIFFERENT MODELS IN DATA-SCARCE SCENARIOS

| Chronos Model | Forecast horizon is 1 h | | | Forecast horizon is 6 h | | | Forecast horizon is 12 h | | | Forecast horizon is 24 h | | | Forecast horizon is 48 h | | |
|---|---|---|---|---|---|---|---|---|---|---|---|---|---|---|---|
| | RMSE (MW) | MAE (MW) | MAPE (%) | RMSE (MW) | MAE (MW) | MAPE (%) | RMSE (MW) | MAE (MW) | MAPE (%) | RMSE (MW) | MAE (MW) | MAPE (%) | RMSE (MW) | MAE (MW) | MAPE (%) |
| With Fine-tuning | 0.79 | 0.56 | 1.36% | 1.59 | 1.10 | 2.68% | 2.08 | 1.49 | 3.66% | 2.48 | 1.82 | 4.42% | 2.93 | 2.14 | 5.26% |
| Without Fine-tuning | 0.75 | 0.52 | 1.29% | 1.51 | 1.03 | 2.54% | 1.97 | 1.39 | 3.47% | 2.35 | 1.70 | 4.19% | 2.78 | 2.00 | 4.99% |

Although the Chronos model already outperforms the baseline methods under the zero-shot load forecasting setting, further performance improvements can be observed after fine-tuning with a small amount of load data, i.e., under the few-shot forecasting setting. This indicates that, while the pre-trained Chronos model possesses strong generalization ability, limited task-specific data can still help the model better adapt to the characteristics of a particular load dataset, leading to reduced forecasting errors.

Moreover, after fine-tuning, the Chronos model consistently achieves lower forecasting errors across different forecast horizons compared to its zero-shot counterpart. This demonstrates that incorporating a small number of historical load samples for fine-tuning can effectively enhance the forecast accuracy of Chronos, especially in practical scenarios where limited data are available. Overall, these results highlight the flexibility of the Chronos model, which can perform well in both zero-shot and few-shot load forecasting settings.

### 4.6 Diebold-Mariano Test

To further explore whether the Chronos model and the baseline models have notable differences in forecast errors, Diebold-Mariano (DM) tests are performed in both data-rich and data-scarce scenarios. The null hypothesis in these tests is that there is no significant difference in forecast error between the Chronos model and a comparison model. According to this null hypothesis, when the absolute value of the DM statistic value is larger than 1.96, the null hypothesis is rejected at a 5% significance level, since the Diebold-Mariano test statistic adheres to a standard normal distribution. Tables VI and VII present the results of the DM tests for data-rich and data-scarce scenarios.

TABLE VI
DM TEST RESULTS BETWEEN CHRONOS MODEL AND BASELINE MODELS IN DATA-RICH SCENARIOS

| Model | Forecast horizon is 1 h | | Forecast horizon is 6 h | | Forecast horizon is 12 h | | Forecast horizon is 24 h | | Forecast horizon is 48 h | |
|---|---|---|---|---|---|---|---|---|---|---|
| | DM statistic value | p-value | DM statistic value | p-value | DM statistic value | p-value | DM statistic value | p-value | DM statistic value | p-value |
| SNM | -21.42 | 0.000 | -10.79 | 0.000 | -8.28 | 0.000 | -6.87 | 0.000 | -6.80 | 0.000 |
| CSBA | -29.54 | 0.000 | -19.11 | 0.000 | -21.24 | 0.000 | -12.99 | 0.000 | -10.69 | 0.000 |
| NPTS | -29.92 | 0.000 | -14.68 | 0.000 | -11.11 | 0.000 | -8.19 | 0.000 | -6.24 | 0.000 |
| AutoETS | -20.53 | 0.000 | -22.91 | 0.000 | -15.25 | 0.000 | -11.76 | 0.000 | -9.97 | 0.000 |
| AutoARIMA | -14.31 | 0.000 | -11.96 | 0.000 | -8.20 | 0.000 | -5.66 | 0.000 | -6.28 | 0.000 |
| TFT | -5.96 | 0.000 | -5.89 | 0.000 | -1.74 | 0.083 | -4.70 | 0.000 | -4.11 | 0.000 |
| DeepAR | -16.89 | 0.000 | -12.80 | 0.000 | -1.61 | 0.107 | -6.09 | 0.000 | -7.45 | 0.000 |
| PatchTST | -6.73 | 0.000 | -10.19 | 0.000 | -10.09 | 0.000 | -5.07 | 0.000 | -3.65 | 0.000 |
| TimeLLM | -28.72 | 0.000 | -21.27 | 0.000 | -20.02 | 0.000 | -18.64 | 0.000 | -15.49 | 0.000 |

TABLE VII
DM TEST RESULTS BETWEEN CHRONOS MODEL AND BASELINE MODELS IN DATA-SCARCE SCENARIOS

| Model | Forecast horizon is 1 h | | Forecast horizon is 6 h | | Forecast horizon is 12 h | | Forecast horizon is 24 h | | Forecast horizon is 48 h | |
|---|---|---|---|---|---|---|---|---|---|---|
| | DM statistics value | p-value | DM statistics value | p-value | DM statistics value | p-value | DM statistics value | p-value | DM statistics value | p-value |
| SNM | -21.42 | 0.000 | -10.79 | 0.000 | -8.28 | 0.000 | -6.87 | 0.000 | -6.80 | 0.000 |
| CSBA | -29.54 | 0.000 | -19.11 | 0.000 | -21.24 | 0.000 | -12.99 | 0.000 | -10.69 | 0.000 |
| NPTS | -28.81 | 0.000 | -14.58 | 0.000 | -11.61 | 0.000 | -8.17 | 0.000 | -6.00 | 0.000 |
| AutoETS | -20.53 | 0.000 | -22.91 | 0.000 | -15.25 | 0.000 | -11.76 | 0.000 | -9.97 | 0.000 |
| AutoARIMA | -9.91 | 0.000 | -11.78 | 0.000 | -8.73 | 0.000 | -7.17 | 0.000 | -5.58 | 0.000 |
| TFT | -3.49 | 0.000 | -11.03 | 0.000 | -8.35 | 0.000 | -6.15 | 0.000 | -5.61 | 0.000 |
| DeepAR | -22.61 | 0.000 | -13.87 | 0.000 | -9.91 | 0.000 | -9.12 | 0.000 | -8.32 | 0.000 |
| PatchTST | -15.92 | 0.000 | -15.64 | 0.000 | -12.66 | 0.000 | -9.40 | 0.000 | -9.31 | 0.000 |
| TimeLLM | -28.82 | 0.000 | -22.58 | 0.000 | -22.46 | 0.000 | -17.81 | 0.000 | -15.35 | 0.000 |

According to the DM test results, in most scenarios, there are significant differences in forecast errors between the Chronos model and baseline models. Specifically, the majority of the DM statistic values are negative, with absolute values far greater than 1.96, indicating that the Chronos model significantly outperforms baseline models across these forecast horizons and scenarios. Additionally, the p-values being 0 further confirm the significance of this difference, meaning the hypothesis of equal forecast errors between the two models can be confidently rejected.

In data-rich scenarios, when the forecast horizon is 12 hours, the DM statistic values for TFT and DeepAR are -1.74 and -1.61, with p-values of 0.083 and 0.107, respectively. These p-values are close to common significance thresholds (e.g., 0.05) but do not reach the level of statistical significance, suggesting that the Chronos model slightly performs better than TFT and DeepAR in these cases.

### 4.7 Probabilistic Forecasting Analytics

In this section, the Chronos model is compared with baseline models for probabilistic load forecasting. Note that TimeLLM is not implemented, because it is only a deterministic forecasting model. As an example, simulations are conducted on the data-rich scenario. The forecast horizons are set at 1, 6, 12, 24, and 48 hours. Each model is executed 30 times, and the average QS and CRPS on the test set are calculated, as presented in Table VIII.

TABLE VIII

QS AND CRPS COMPARISON OF CHRONOS AND BASELINES

| Model | Forecast horizon is 1 h | | Forecast horizon is 6 h | | Forecast horizon is 12 h | | Forecast horizon is 24 h | | Forecast horizon is 48 h | |
|---|---|---|---|---|---|---|---|---|---|---|
| | QS (MW) | CRPS (MW) | QS (MW) | CRPS (MW) | QS (MW) | CRPS (MW) | QS (MW) | CRPS (MW) | QS (MW) | CRPS (MW) |
| SNM | 1.14 | 2.11 | 1.14 | 2.11 | 1.14 | 2.11 | 1.14 | 2.11 | 1.32 | 2.43 |
| CSBA | 1.27 | 2.87 | 1.56 | 3.08 | 1.72 | 3.45 | 1.45 | 2.79 | 1.58 | 3.03 |
| NPTS | 1.55 | 3.26 | 1.55 | 3.26 | 1.56 | 3.27 | 1.56 | 3.28 | 1.57 | 3.31 |
| AutoETS | 0.36 | 0.65 | 1.11 | 2.14 | 1.08 | 1.98 | 1.35 | 2.59 | 1.50 | 2.78 |
| AutoARIMA | 0.31 | 0.57 | 0.86 | 1.67 | 1.06 | 2.07 | 1.30 | 2.56 | 1.40 | 2.76 |
| TFT | 0.24 | 0.45 | 0.53 | 1.02 | 0.75 | 1.46 | 0.92 | 1.63 | 1.22 | 2.38 |
| DeepAR | 0.39 | 0.69 | 0.81 | 1.72 | 0.76 | 1.34 | 1.20 | 2.52 | 3.85 | 7.95 |
| PatchTST | 0.24 | 0.44 | 0.61 | 1.20 | 0.73 | 1.42 | 0.94 | 1.68 | 1.08 | 2.06 |
| Chronos | **0.23** | **0.40** | **0.44** | **0.82** | **0.60** | **1.11** | **0.71** | **1.31** | **0.85** | **1.58** |

TABLE IX

PICP, ACE, PIAW, AND WS COMPARISON OF CHRONOS AND BASELINES

| Model | 95% PINC | | | | 90% PINC | | | | 80% PINC | | | | 70% PINC | | | |
|---|---|---|---|---|---|---|---|---|---|---|---|---|---|---|---|---|
| | PICP (%) | ACE (%) | PIAW (MW) | WS (MW) | PICP (%) | ACE (%) | PIAW (MW) | WS (MW) | PICP (%) | ACE (%) | PIAW (MW) | WS (MW) | PICP (%) | ACE (%) | PIAW (MW) | WS (MW) |
| SNM | 94.23% | -0.77% | 15.31 | 19.81 | 89.31% | -0.69% | 12.85 | 17.02 | 81.69% | 1.68% | 10.01 | 14.05 | 74.45% | 4.45% | 8.10 | 12.14 |
| CSBA | 93.57% | -1.43% | 17.66 | 21.67 | 89.19% | -0.81% | 15.46 | 19.22 | 79.44% | -0.56% | 12.21 | 16.55 | 69.41% | -0.59% | 9.90 | 14.68 |
| NPTS | 90.11% | -4.89% | 16.56 | 22.07 | 82.07% | -7.93% | 14.17 | 19.67 | 71.25% | -8.75% | 11.60 | 17.05 | 62.34% | -7.66% | 9.63 | 15.42 |
| AutoETS | 91.25% | -3.75% | 15.80 | 20.53 | 84.89% | -5.11% | 13.26 | 18.11 | 72.44% | -7.56% | 10.33 | 15.36 | 60.99% | -9.01% | 8.36 | 13.65 |
| AutoARIMA | 74.04% | -20.96% | 8.86 | 31.55 | 66.76% | -23.24% | 7.44 | 23.09 | 55.88% | -24.12% | 5.80 | 16.87 | 47.66% | -22.34% | 4.69 | 13.88 |
| TFT | 88.55% | -6.45% | 9.21 | 14.82 | 76.51% | -13.49% | 7.11 | 13.30 | 63.51% | -16.49% | 5.31 | 10.88 | 53.43% | -16.57% | 4.19 | 9.21 |
| DeepAR | 82.69% | -12.31% | 9.42 | 22.96 | 76.24% | -13.76% | 7.73 | 17.78 | 66.74% | -13.26% | 5.96 | 13.37 | 58.61% | -11.39% | 4.79 | 11.14 |
| PatchTST | 94.05% | -0.95% | 13.07 | 16.45 | 87.41% | -2.59% | 10.66 | 14.42 | 75.53% | -4.47% | 8.09 | 12.20 | 65.57% | -4.43% | 6.46 | 10.72 |
| Chronos | 87.87% | -7.13% | 7.63 | 13.29 | 81.91% | -8.09% | 6.33 | 10.84 | 72.48% | -7.52% | 4.96 | 8.51 | 62.98% | -7.02% | 3.99 | 7.30 |

***1) Comprehensive Comparison with QS and CRPS***

As an LLM-based forecasting model, the Chronos model significantly outperforms baseline models for probabilistic load forecasting with various forecast horizons. For example, in the 24-hour forecast, the Chronos model shows considerable reductions in QS of 37.72%, 51.03%, 54.49%, 47.41%, 45.38%, 22.83%, 40.83%, and 24.47% compared to SNM, CSBA, NPTS, AutoETS, AutoARIMA, TFT, DeepAR, and PatchTST, respectively. Similarly, Chronos achieves reductions in CRPS of 37.91%, 53.05%, 60.06%, 49.42%, 48.83%, 19.63%, 48.02%, and 22.02% compared to the same models. These results quantitatively highlight the advantages of the Chronos model in probabilistic load forecasting.

2) ***Comparison Under different PINC levels***

In addition to QS and CRPS, Table IX presents the average PICP, ACE, PIAW, and WS under different PINC levels when the forecast horizon is 24 hours.

At different PINC levels, the Chronos model shows a significantly higher PICP compared to AutoARIMA and DeepAR, while its PIAW is also lower than these two models. This indicates that the Chronos model ensures a high coverage of its forecasts while providing more accurate PIs, demonstrating its superiority in capturing load variations.

Although PatchTST exhibits a higher PICP than Chronos at certain PINC levels, its PIAW is also significantly larger. This indicates that PatchTST produces wider prediction intervals, which may lead to increased reserve requirements in practical grid operations. In general, an appropriate balance between PICP and PIAW is desirable: high PICP ensures safety by covering more possible outcomes, while narrower PIAW reduces unnecessary operational reserves. Since PICP and PIAW alone cannot fully determine which model is preferable, the WS metric serves as a comprehensive measure, combining both coverage and interval width. As shown in Table IX, Chronos achieves the lowest WS across all PINC levels, demonstrating that it effectively balances coverage and interval width. This implies that Chronos provides accurate probabilistic forecasts that maintain grid reliability while minimizing additional reserve costs, offering an optimal trade-off between safety and operational efficiency.

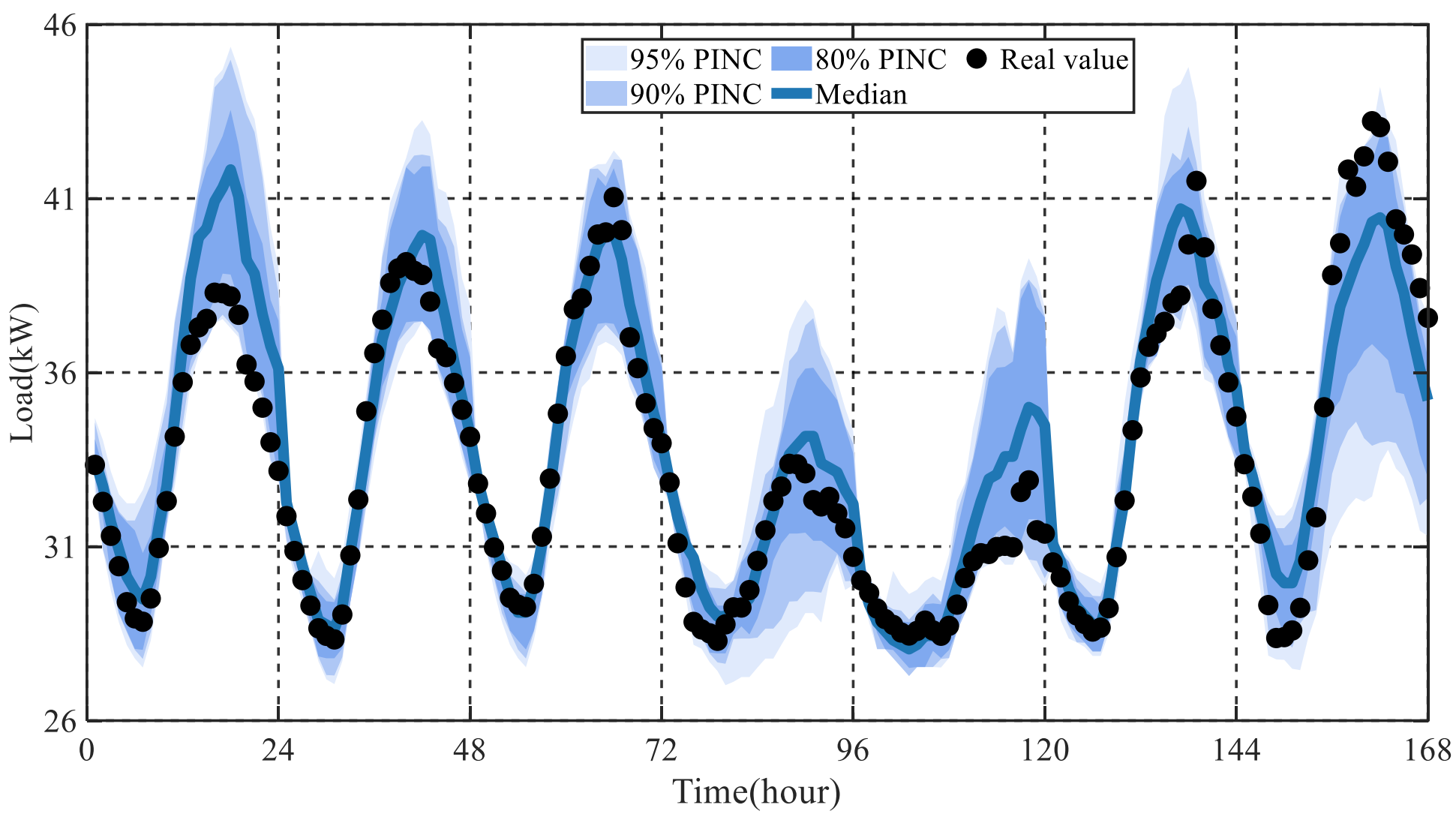

Fig. 5. Probabilistic Forecasting Results of Chronos over a 7-Day Period for 24-Hour Forecast Horizon at 95%, 90%, and 80% PINC Levels.

***3) Visual analysis for probabilistic forecasting***

To visualize the performance of Chronos in probabilistic forecasting, Fig. 6 shows the forecast results over one week (from 8 February 2012 to 14 February 2012) in a data-rich scenario with a forecast horizon of 24 hours. The blue shaded areas represent probabilistic forecasts at different PINC levels, while the black dots mark the actual load values**.**

It is obvious that the PIs generated by Chronos effectively capture the actual load across various PINC levels. Even though Chronos is not trained on this specific dataset, it exhibits strong generalization in probabilistic load forecasting, highlighting its robustness and adaptability for accurate load forecasting in previously unseen scenarios.

### *4.8 Time Complexity Comparison*

In this section, the time complexity of each model is discussed. As a simple example, load forecasting is performed in the data-rich scenario with a forecast horizon of 24 hours. Each model is executed 30 times, and the average calculation time and inference time are calculated, as presented in Table X. Note that the training time of TimeLLM is not mentioned in [20], so it is not discussed.

TABLE X
CALCULATION TIME AND INFERENCE TIME OF EACH MODEL

| Model | Training time | Inference time |
|---|---|---|
| SNM | 7.58 s | 0.05 s |
| CSBA | 39.12 s | 1.47 s |
| NPTS | 1.12 s | 0.17 s |
| AutoETS | 59.90 s | 12.91 s |
| AutoARIMA | 131.98 s | 16.06 s |
| TFT | 653.23 s | 0.09 s |
| DeepAR | 558.31 s | 0.13 s |
| PatchTST | 246.79 s | 0.04 s |
| TimeLLM | none | 3.74 s |
| Chronos | 63.05 h | 2.76 s |

The training and inference times for statistical models are relatively short, demonstrating high computational efficiency. For example, the training time and inference time of AutoETS are 59.90 seconds and 12.91 seconds, respectively. Meanwhile, SNM and NPTS have training times of 7.58 seconds and 1.12 seconds respectively, with inference times of only 0.05 seconds and 0.17 seconds. This indicates that statistical models in load forecasting tasks can produce results quickly with minimal computational cost.

In contrast, deep learning models have significantly longer training times. For example, TFT and DeepAR require more than 500 seconds for model training, while PatchTST has a relatively shorter training time of 246.79 seconds. However, these deep learning models show high efficiency in the inference stage, with inference times of less than 0.15 seconds, indicating their ability to conduct rapid forecasting once trained.

The Chronos model requires a significantly longer training time (63.05 hours) compared to statistical models, which can be trained in seconds. However, it is important to note that this training is a one-time, offline pre-training process on large-scale time series datasets. Once trained, the Chronos model can perform inference in only 2.76 seconds, which is sufficiently fast for real-time or operational load forecasting tasks. In contrast, while statistical models such as AutoETS or SNM have extremely short training times, their forecasting accuracy is much lower, as shown in Sections 4.2–4.5. Achieving both high accuracy and rapid training on a small dataset is inherently challenging, and currently no model can achieve simultaneously minimal training time, low inference latency, and high forecast accuracy across diverse scenarios.

Therefore, the computational cost of pre-training Chronos is justified by its superior forecasting performance, robustness to data scarcity, and rapid inference capability, making it highly suitable for practical grid operations, where accurate predictions are critical and inference speed, rather than training speed, is the key factor.

## 5 Case Study II

### *5.1 Simulation Settings*

To discuss the generalization of each model, four real-world load datasets in [25] are used to conduct simulations and analyses. Note that these four datasets have not been used for parameterizing the LLMs (e.g., Chronos model and TimeLLM).

Midea Dataset: This is an hourly load dataset, which is collected from Midea Group, an electrical appliance manufacturer. The time spans from 1 May 2017 to 20 August 2017. The training and validation sets cover the period from May 1 to May 31 (30 days, with 80% for training and 20% for validation). The test set spans from June 1 to August 20 (81 days).

Nongfu Dataset: This is an hourly load dataset, which is collected from Nongfu Spring, a packaging water supplier. The time spans from 1 May 2017 to 20 August 2017. The training and validation sets cover the period from May 1 to May 31 (30 days, with 80% for training and 20% for validation). The test set spans from June 1 to August 20 (81 days).

Two datasets from mathematics competition: These two datasets come from different regions. Specifically, the first dataset covers the period from January 1, 2014, to December 31, 2014, while the second dataset spans from January 1, 2013, to December 31, 2013. Both datasets have an hourly time resolution. For both datasets, the training and validation sets cover the period from January 1 to June 30 (6 months, with 80% for training and 20% for validation). The test set spans from July 1 to December 31 (6 months).

### *5.2 Results and Analyses*

To ensure a fair comparison, the parameters of each model are also determined by Bayesian optimization. The forecast horizons are set at 1, 6, 12, 24, and 48 hours. Each model is executed 30 times, and the average test set error for deterministic forecasting is calculated, as presented in Tables XI-XV.

The results show that the Chronos model outperforms the baseline models across various datasets and forecast horizons (e.g., 1 hour, 6 hours, 12 hours, 24 hours, and 48 hours) for deterministic load forecasting. For example, in the Nongfu dataset with a forecast horizon of 12 hours, the Chronos model achieves RMSE, MAE, and MAPE of 1248.77 kW, 939.24 kW, and 6.41%, respectively. Compared to baseline models, Chronos model reduces the RMSE, MAE, and MAPE by approximately 21.43%-57.04%, 23.12%-84.71%, and 22.90%-83.50%, respectively.

In addition to deterministic load forecasting, Table XV presents the average QS and CRPS of each model for probabilistic load forecasting with a forecast horizon of 24 hours.

In the simulation results across these four datasets, Chronos model shows significantly lower QS and CRPS compared to baseline models, demonstrating its superiority over baseline models in probabilistic load forecasting and strong generalization capability. For instance, in the Midea dataset, Chronos reduces QS by 44.82%, 63.71%, 44.49%, 66.14%, 41.38%, 37.67%, 71.60%, and 35.00% compared to SNM, CSBA, NPTS, AutoETS, AutoARIMA, TFT, DeepAR, and PatchTST, respectively. Similarly, Chronos model lowers CRPS by 39.63%, 63.19%, 38.10%, 66.64%, 39.84%, 34.99%, 74.10%, and 34.47% against the same models.

TABLE XI
PERFORMANCE EVALUATION OF DIFFERENT MODELS ON THE MIDEA DATASET

| Model | Forecast horizon is 1 h | | | Forecast horizon is 6 h | | | Forecast horizon is 12 h | | | Forecast horizon is 24 h | | | Forecast horizon is 48 h | | |
|---|---|---|---|---|---|---|---|---|---|---|---|---|---|---|---|
| | RMSE (kW) | MAE (kW) | MAPE (%) | RMSE (kW) | MAE (kW) | MAPE (%) | RMSE (kW) | MAE (kW) | MAPE (%) | RMSE (kW) | MAE (kW) | MAPE (%) | RMSE (kW) | MAE (kW) | MAPE (%) |
| SNM | 2159.44 | 1340.61 | 29.23% | 2159.44 | 1340.61 | 29.23% | 2159.44 | 1340.61 | 29.23% | 2159.44 | 1340.61 | 29.23% | 2340.88 | 1537.10 | 36.86% |
| CSBA | 2531.61 | 2187.36 | 32.54% | 2961.10 | 2564.12 | 39.22% | 2931.08 | 2619.66 | 41.99% | 2944.41 | 2527.06 | 45.05% | 3099.70 | 2666.14 | 54.11% |
| NPTS | 2417.34 | 1494.25 | 51.58% | 2427.44 | 1505.65 | 51.78% | 2427.44 | 1510.51 | 51.88% | 2418.91 | 1498.67 | 51.76% | 2435.48 | 1519.38 | 52.02% |
| AutoETS | 1059.97 | 719.70 | 9.30% | 2650.73 | 1940.87 | 26.00% | 2406.79 | 1820.59 | 28.24% | 3229.77 | 2675.06 | 36.09% | 3321.98 | 2799.38 | 42.24% |
| AutoARIMA | 688.27 | 454.05 | 8.63% | 1647.88 | 1047.44 | 21.35% | 2309.49 | 1309.25 | 28.12% | 2037.36 | 1359.32 | 30.52% | 2316.10 | 1576.87 | 40.91% |
| TFT | 620.82 | 436.23 | 7.69% | 1199.97 | 809.20 | 17.90% | 1443.98 | 1022.92 | 25.39% | 2324.59 | 1702.41 | 43.88% | 2100.44 | 1391.27 | 41.43% |
| DeepAR | 2517.18 | 2139.02 | 32.59% | 4324.83 | 3656.02 | 57.29% | 2361.10 | 1707.59 | 37.87% | 4560.12 | 3987.77 | 51.31% | 3602.29 | 3184.41 | 49.69% |
| PatchTST | 898.82 | 695.33 | 12.71% | 1238.78 | 834.92 | 18.44% | 1613.03 | 1066.51 | 26.38% | 1908.88 | 1275.02 | 32.33% | 2233.29 | 1685.78 | 40.09% |
| TimeLLM | 3376.43 | 2672.60 | 57.63% | 5983.88 | 4913.26 | 83.38% | 3916.26 | 3113.27 | 62.36% | 5040.53 | 4254.96 | 77.26% | 4260.13 | 3544.32 | 69.25% |
| Chronos | **412.76** | **246.57** | **3.74%** | **780.37** | **460.87** | **8.45%** | **1029.42** | **569.70** | **10.79%** | **1494.05** | **811.86** | **18.53%** | **1845.69** | **1007.63** | **27.14%** |

TABLE XII
PERFORMANCE EVALUATION OF DIFFERENT MODELS ON THE NONGFU DATASET

| Model | Forecast horizon is 1 h | | | Forecast horizon is 6 h | | | Forecast horizon is 12 h | | | Forecast horizon is 24 h | | | Forecast horizon is 48 h | | |
|---|---|---|---|---|---|---|---|---|---|---|---|---|---|---|---|
| | RMSE (kW) | MAE (kW) | MAPE (%) | RMSE (kW) | MAE (kW) | MAPE (%) | RMSE (kW) | MAE (kW) | MAPE (%) | RMSE (kW) | MAE (kW) | MAPE (%) | RMSE (kW) | MAE (kW) | MAPE (%) |
| SNM | 2079.68 | 1577.91 | 10.49% | 2079.68 | 1577.91 | 10.49% | 2079.68 | 1577.91 | 10.49% | 2079.68 | 1577.91 | 10.49% | 2124.13 | 1643.27 | 10.91% |
| CSBA | 1666.70 | 1389.97 | 8.95% | 1894.37 | 1558.79 | 10.08% | 1915.80 | 1558.82 | 10.04% | 2100.90 | 1676.13 | 10.88% | 2137.70 | 1734.85 | 11.25% |
| NPTS | 2219.12 | 1812.19 | 12.06% | 2229.42 | 1819.27 | 12.09% | 2215.72 | 1804.34 | 11.99% | 2232.26 | 1826.43 | 12.14% | 2257.31 | 1848.91 | 12.29% |
| AutoETS | 693.64 | 513.04 | 3.44% | 1509.99 | 1076.14 | 7.40% | 2039.41 | 1513.93 | 10.10% | 2105.48 | 1558.67 | 10.95% | 2168.19 | 1650.36 | 11.51% |
| AutoARIMA | 835.28 | 613.51 | 4.12% | 1597.59 | 1163.60 | 7.80% | 1856.60 | 1374.92 | 9.12% | 1953.82 | 1475.95 | 9.95% | 2002.35 | 1575.02 | 10.54% |
| TFT | 758.58 | 573.95 | 3.86% | 1529.27 | 1115.33 | 7.68% | 1589.42 | 1221.77 | 8.32% | 2217.96 | 1723.66 | 11.89% | 2408.89 | 1866.96 | 12.76% |
| DeepAR | 1701.70 | 1421.39 | 9.55% | 3717.84 | 2859.88 | 19.90% | 6596.20 | 6142.06 | 38.86% | 6473.20 | 5914.23 | 37.77% | 6350.84 | 5915.19 | 37.51% |
| PatchTST | 729.63 | 555.93 | 3.79% | 1425.90 | 1084.89 | 7.52% | 2120.45 | 1675.31 | 11.37% | 1955.48 | 1565.59 | 10.59% | 1984.38 | 1583.86 | 10.59% |
| TimeLLM | 2979.68 | 2392.67 | 14.83% | 2241.70 | 1750.86 | 11.94% | 2906.79 | 2251.74 | 15.25% | 2653.25 | 2032.53 | 13.63% | 4155.91 | 3179.87 | 20.95% |
| Chronos | **636.33** | **474.60** | **3.21%** | **1022.03** | **771.52** | **5.27%** | **1248.77** | **939.24** | **6.41%** | **1649.74** | **1214.95** | **8.31%** | **1709.78** | **1333.84** | **9.02%** |

TABLE XIII
PERFORMANCE EVALUATION OF DIFFERENT MODELS ON THE FIRST COMPETITION DATASET

| Model | Forecast horizon is 1 h | | | Forecast horizon is 6 h | | | Forecast horizon is 12 h | | | Forecast horizon is 24 h | | | Forecast horizon is 48 h | | |
|---|---|---|---|---|---|---|---|---|---|---|---|---|---|---|---|
| | RMSE (MW) | MAE (MW) | MAPE (%) | RMSE (MW) | MAE (MW) | MAPE (%) | RMSE (MW) | MAE (MW) | MAPE (%) | RMSE (MW) | MAE (MW) | MAPE (%) | RMSE (MW) | MAE (MW) | MAPE (%) |
| SNM | 1049.72 | 650.02 | 8.73% | 1049.72 | 650.02 | 8.73% | 1049.72 | 650.02 | 8.73% | 614.88 | 367.56 | 4.86% | 1235.83 | 774.17 | 10.55% |
| CSBA | 1627.48 | 1366.55 | 17.33% | 1833.01 | 1535.21 | 19.85% | 1827.89 | 1604.48 | 21.27% | 1763.26 | 1523.46 | 20.73% | 1871.85 | 1601.28 | 22.00% |
| NPTS | 1256.53 | 891.14 | 12.84% | 1255.24 | 888.19 | 12.80% | 1258.03 | 891.55 | 12.84% | 1254.64 | 888.24 | 12.80% | 1270.08 | 903.60 | 13.00% |
| AutoETS | 500.30 | 309.15 | 4.27% | 621.13 | 384.47 | 5.30% | 690.64 | 437.50 | 6.03% | 954.42 | 605.18 | 8.18% | 1144.52 | 737.61 | 10.17% |
| AutoARIMA | 296.40 | 172.48 | 2.29% | 555.52 | 311.13 | 4.05% | 636.33 | 389.68 | 5.11% | 967.70 | 618.84 | 8.03% | 750.84 | 486.71 | 6.55% |
| TFT | 214.03 | 154.15 | 2.03% | 543.42 | 332.17 | 4.52% | 536.55 | 371.41 | 5.16% | 770.19 | 500.04 | 6.48% | 750.84 | 486.71 | 6.55% |
| DeepAR | 299.83 | 224.36 | 3.12% | 461.15 | 328.12 | 4.43% | 556.49 | 402.81 | 6.12% | 937.85 | 701.69 | 10.34% | 1213.82 | 918.82 | 13.51% |
| PatchTST | 217.55 | 149.48 | 1.99% | 537.36 | 327.59 | 4.10% | 527.08 | 303.31 | 3.85% | 614.88 | 367.56 | 4.86% | 906.87 | 578.71 | 7.77% |
| TimeLLM | 1800.78 | 1539.08 | 20.14% | 3435.89 | 2895.89 | 41.28% | 2839.64 | 2203.05 | 30.99% | 2721.32 | 2323.77 | 32.52% | 2733.83 | 2241.51 | 30.93% |
| Chronos | **116.99** | **87.57** | **1.17%** | **302.74** | **145.88** | **1.97%** | **355.48** | **173.57** | **2.39%** | **588.67** | **264.27** | **3.67%** | **689.07** | **322.51** | **4.58%** |

TABLE XIV
PERFORMANCE EVALUATION OF DIFFERENT MODELS ON THE SECOND COMPETITION DATASET

| Model | Forecast horizon is 1 h | | | Forecast horizon is 6 h | | | Forecast horizon is 12 h | | | Forecast horizon is 24 h | | | Forecast horizon is 48 h | | |
|---|---|---|---|---|---|---|---|---|---|---|---|---|---|---|---|
| | RMSE (MW) | MAE (MW) | MAPE (%) | RMSE (MW) | MAE (MW) | MAPE (%) | RMSE (MW) | MAE (MW) | MAPE (%) | RMSE (MW) | MAE (MW) | MAPE (%) | RMSE (MW) | MAE (MW) | MAPE (%) |
| SNM | 681.47 | 469.13 | 5.98% | 681.47 | 469.13 | 5.98% | 681.47 | 469.13 | 5.98% | 681.47 | 469.13 | 5.98% | 787.44 | 560.53 | 7.20% |
| CSBA | 1604.79 | 1437.36 | 19.47% | 1842.37 | 1641.54 | 22.63% | 1947.90 | 1800.43 | 25.24% | 1662.30 | 1420.21 | 20.92% | 1698.99 | 1447.67 | 21.30% |
| NPTS | 1078.32 | 827.05 | 11.39% | 1069.39 | 820.29 | 11.31% | 1070.63 | 822.11 | 11.32% | 1074.23 | 824.21 | 11.35% | 1083.71 | 835.10 | 11.50% |
| AutoETS | 347.92 | 243.36 | 3.20% | 446.05 | 310.38 | 4.09% | 476.18 | 332.86 | 4.37% | 627.07 | 440.61 | 5.62% | 724.66 | 517.60 | 6.63% |
| AutoARIMA | 203.40 | 148.61 | 1.97% | 704.53 | 403.43 | 5.17% | 1080.14 | 515.79 | 6.59% | 859.92 | 546.92 | 6.96% | 959.80 | 630.68 | 8.08% |
| TFT | 285.58 | 223.16 | 2.88% | 565.79 | 410.04 | 5.34% | 564.79 | 431.09 | 5.87% | 838.89 | 635.77 | 8.19% | 803.25 | 595.20 | 7.79% |
| DeepAR | 276.90 | 215.77 | 3.10% | 381.77 | 294.70 | 4.19% | 636.45 | 519.58 | 7.35% | 1334.16 | 1107.25 | 15.36% | 1856.09 | 1470.68 | 20.95% |
| PatchTST | 314.18 | 260.30 | 3.54% | 400.25 | 265.20 | 3.41% | 388.60 | 235.76 | 2.97% | 608.69 | 395.58 | 4.91% | 667.78 | 476.46 | 5.96% |
| TimeLLM | 2349.43 | 1962.26 | 23.34% | 2262.21 | 1756.26 | 26.01% | 3218.90 | 2581.81 | 38.27% | 3337.37 | 2867.16 | 40.51% | 2270.25 | 1887.29 | 27.55% |
| Chronos | **131.74** | **102.29** | **1.38%** | **263.41** | **169.66** | **2.24%** | **303.96** | **193.61** | **2.55%** | **432.80** | **269.48** | **3.44%** | **555.86** | **351.98** | **4.45%** |

TABLE XV
THE RESULTS OF PROBABILISTIC LOAD FORECASTING IN FOUR DATASETS

| Model | Midea Dataset | | Nongfu Dataset | | First competition dataset | | Second competition dataset | |
|---|---|---|---|---|---|---|---|---|
| | QS (kW) | CRPS (kW) | QS (kW) | CRPS (kW) | QS (MW) | CRPS (MW) | QS (MW) | CRPS (MW) |
| SNM | 613.81 | 1094.85 | 626.39 | 1140.73 | 301.74 | 531.89 | 198.98 | 351.09 |
| CSBA | 933.23 | 1795.55 | 644.35 | 1272.64 | 556.55 | 1066.41 | 530.13 | 978.73 |
| NPTS | 610.15 | 1067.91 | 710.54 | 1363.35 | 350.43 | 605.67 | 320.55 | 560.94 |
| AutoETS | 1000.11 | 1981.68 | 623.79 | 1034.51 | 261.69 | 487.34 | 179.99 | 336.04 |
| AutoARIMA | 577.75 | 1098.78 | 621.28 | 1166.15 | 282.94 | 488.71 | 260.34 | 466.33 |
| TFT | 543.37 | 1016.78 | 694.70 | 1280.55 | 225.57 | 445.59 | 204.15 | 366.93 |
| DeepAR | 1192.59 | 2552.46 | 1054.01 | 2184.67 | 327.16 | 628.79 | 456.22 | 879.52 |
| PatchTST | 521.06 | 1008.69 | 597.06 | 1017.49 | 149.80 | 282.09 | 180.61 | 349.33 |
| Chronos | **338.67** | **661.01** | **476.03** | **884.23** | **113.21** | **216.62** | **107.72** | **198.59** |

## 6 CONCLUSION

Inspired by the great success of pre-trained LLMs, this paper proposes a zero-shot load forecasting approach by adopting LLMs (i.e., Chronos model). By utilizing its extensive pre-trained knowledge, the Chronos model enables accurate load forecasting in data-scarce scenarios. The comprehensive simulations and analyses in five real-world datasets result in the following conclusions:

The Chronos model significantly outperforms nine customized baseline models for both deterministic load forecasting and probabilistic load forecasting, with various forecast horizons (e.g., 1 hour, 6 hours, 12 hours, 24 hours and 48 hours). Compared to baseline models, Chronos reduces the RMSE, MAE, and MAPE by approximately 7.34%-84.30%, 8.96%-86.21%, and 7.87%-85.60%, respectively. Similarly, Chronos achieves reductions in CRPS of 19.63%-60.06% and QS of 22.83%-54.49% compared to the baseline models. This confirms the superiority and flexibility of the Chronos model, making it an effective solution for both deterministic and probabilistic load forecasting in data-scarce scenarios.

Chronos model uses the LLM framework but is directly pre-trained on massive time series data rather than text data. This direct training on time series data gives Chronos a significant advantage in handling time series analysis tasks, resulting in superior performance load forecasting compared to other LLMs, such as TimeLLM, which are trained on text data.

Although the Chronos model has significantly longer training times (e.g. tens of hours) than statistical and deep learning models, its inference takes only a few seconds, making it feasible for real-world forecasting tasks.

## Funding

This work is funded by the Swiss Federal Office of Energy (Grant No. SI/502135–01). Also, this work is carried out in the frame of the “UrbanTwin: An urban digital twin for climate action: Assessing policies and solutions for energy, water and infrastructure” project with the financial support of the ETH-Domain Joint Initiative program in the Strategic Area Energy, Climate and Sustainable Environment.